\documentclass[english,smallextended]{svjour3}
\usepackage[latin9]{inputenc}
\usepackage{babel}
\usepackage{verbatim}
\usepackage{amsmath}
\usepackage{amssymb}
\usepackage{graphicx}
\usepackage[unicode=true]
 {hyperref}

\makeatletter

\providecommand{\tabularnewline}{\\}

\smartqed

\usepackage{graphicx}

\institute{
Tu Dinh Nguyen, Truyen Tran, Dinh Phung, Svetha Venkatesh \at  Center for Pattern Recognition and Data Analytics, Deakin University, Australia.\\
Corresponding \email{tu.nguyen@deakin.edu.au}.\\
} 

\@ifundefined{showcaptionsetup}{}{%
 \PassOptionsToPackage{caption=false}{subfig}}
\usepackage{subfig}
\makeatother

\begin{document}

\global\long\def\argmax#1{\underset{_{#1}}{\text{argmax}} }

\global\long\def\argmin#1{\underset{_{#1}}{\text{argmin}} }

\global\long\def\likelihood{\mathcal{L}}

\global\long\def\loglik{\mathcal{\ell}}

\global\long\def\tr{\mathrm{Tr}}

\global\long\def\bigcdot{\raisebox{-0.5ex}{\scalebox{1.5}{\ensuremath{\cdot}}}}

\global\long\def\grad{\bigtriangleup}

\global\long\def\cov{\textnormal{Cov}}

\global\long\def\var{\mathbb{V}}

\global\long\def\gv{\mid}

\global\long\def\bigO{\mathcal{O}}

\global\long\def\sig{\textrm{sig}}

\global\long\def\ci{\perp}

\global\long\def\natureset{\mathbb{N}}

\global\long\def\realset{\mathbb{R}}

\global\long\def\expect{\mathbb{E}}

\global\long\def\ent{\mathcal{H}}

\global\long\def\realn{\realset^{n}}

\global\long\def\integerset{\mathbb{Z}}

\global\long\def\natset{\integerset}

\global\long\def\interger{\integerset}

\global\long\def\natn{\natset^{n}}

\global\long\def\partf{\mathcal{Z}}

\global\long\def\rational{\mathbb{Q}}

\global\long\def\realPlusn{\mathbb{R_{+}^{n}}}

\global\long\def\comp{\complexset}
 \global\long\def\complexset{\mathbb{C}}

\global\long\def\dataset{\mathcal{D}}

\global\long\def\class{\mathcal{C}}

\global\long\def\normal{\mathcal{N}}

\global\long\def\nnset{\mathsf{N}}

\global\long\def\a{\mathrm{a}}

\global\long\def\ba{\mathbf{a}}

\global\long\def\bA{\mathbf{A}}

\global\long\def\seta{\mathcal{A}}

\global\long\def\ta{\tilde{\mathrm{a}}}

\global\long\def\bta{\mathbf{\tilde{a}}}

\global\long\def\b{\mathrm{b}}

\global\long\def\bb{\mathbf{b}}

\global\long\def\bB{\mathbf{B}}

\global\long\def\setb{\mathcal{B}}

\global\long\def\c{\mathrm{c}}

\global\long\def\C{\mathrm{C}}

\global\long\def\bC{\mathbf{C}}

\global\long\def\setc{\mathcal{C}}

\global\long\def\d{\mathrm{d}}

\global\long\def\bd{\mathbf{d}}

\global\long\def\D{\mathrm{D}}

\global\long\def\bD{\mathbf{D}}

\global\long\def\e{\mathrm{e}}

\global\long\def\be{\mathbf{e}}

\global\long\def\E{\mathrm{E}}

\global\long\def\bE{\mathbf{E}}

\global\long\def\sete{\mathcal{E}}

\global\long\def\f{\mathrm{f}}

\global\long\def\bf{\mathbf{f}}

\global\long\def\bF{\mathbf{F}}

\global\long\def\g{\mathrm{g}}

\global\long\def\G{\mathrm{G}}

\global\long\def\bg{\mathbf{g}}

\global\long\def\bG{\mathbf{G}}

\global\long\def\setg{\mathcal{G}}

\global\long\def\h{\mathrm{h}}

\global\long\def\bh{\mathbf{h}}

\global\long\def\H{\mathrm{H}}

\global\long\def\bH{\mathbf{H}}

\global\long\def\hh{\hat{\h}}

\global\long\def\bhh{\hat{\mathbf{h}}}

\global\long\def\seth{\mathcal{H}}

\global\long\def\bI{\mathbf{I}}

\global\long\def\I{\mathrm{I}}

\global\long\def\K{\mathrm{K}}

\global\long\def\bK{\mathbf{K}}

\global\long\def\setk{\mathcal{K}}

\global\long\def\L{\mathrm{L}}

\global\long\def\m{\mathrm{m}}

\global\long\def\bm{\mathbf{m}}

\global\long\def\M{\mathrm{M}}

\global\long\def\barm{\bar{\mathrm{m}}}

\global\long\def\N{\mathrm{N}}

\global\long\def\barn{\bar{\mathrm{n}}}

\global\long\def\p{\mathrm{p}}

\global\long\def\bp{\mathbf{p}}

\global\long\def\bP{\mathbf{P}}

\global\long\def\q{\mathrm{q}}

\global\long\def\tq{\tilde{\mathrm{q}}}

\global\long\def\barq{\bar{\mathrm{q}}}

\global\long\def\bq{\mathbf{q}}

\global\long\def\br{\mathbf{r}}

\global\long\def\R{\mathrm{R}}

\global\long\def\s{\mathrm{s}}

\global\long\def\bs{\mathbf{s}}

\global\long\def\S{\mathrm{S}}

\global\long\def\bS{\mathbf{S}}

\global\long\def\sets{\mathcal{S}}

\global\long\def\t{\mathrm{t}}

\global\long\def\bt{\mathbf{t}}

\global\long\def\T{\mathrm{T}}

\global\long\def\bT{\mathbf{T}}

\global\long\def\u{\mathrm{u}}

\global\long\def\bu{\mathbf{u}}

\global\long\def\bU{\mathbf{U}}

\global\long\def\v{\mathrm{v}}

\global\long\def\bv{\mathbf{v}}

\global\long\def\V{\mathrm{V}}

\global\long\def\bV{\mathbf{V}}

\global\long\def\tv{\tilde{\mathrm{v}}}

\global\long\def\btv{\tilde{\mathbf{v}}}

\global\long\def\setv{\mathcal{V}}

\global\long\def\w{\mathrm{w}}

\global\long\def\bw{\mathbf{w}}

\global\long\def\W{\mathrm{W}}

\global\long\def\bW{\mathbf{W}}

\global\long\def\barw{\bar{\w}}

\global\long\def\bbarw{\bar{\bw}}

\global\long\def\hw{\hat{\w}}

\global\long\def\bhw{\hat{\mathbf{w}}}

\global\long\def\tW{\tilde{\mathrm{W}}}

\global\long\def\btW{\mathbf{\tilde{W}}}

\global\long\def\x{\mathrm{x}}

\global\long\def\bx{\mathbf{x}}

\global\long\def\tx{\tilde{\mathrm{x}}}

\global\long\def\btx{\mathbf{\tilde{x}}}

\global\long\def\hx{\hat{\x}}

\global\long\def\bhx{\hat{\mathbf{\bx}}}

\global\long\def\X{\mathrm{X}}

\global\long\def\bX{\mathbf{X}}

\global\long\def\bhX{\hat{\mathbf{X}}}

\global\long\def\setx{\mathcal{X}}

\global\long\def\y{\mathrm{y}}

\global\long\def\Y{\mathrm{Y}}

\global\long\def\by{\mathbf{y}}

\global\long\def\bY{\mathbf{Y}}

\global\long\def\ty{\tilde{y}}

\global\long\def\bary{\bar{y}}

\global\long\def\bbary{\mathbf{\bar{y}}}

\global\long\def\z{\mathrm{z}}

\global\long\def\bz{\mathbf{z}}

\global\long\def\Z{\mathrm{Z}}

\global\long\def\bZ{\mathbf{Z}}

\global\long\def\bzero{\boldsymbol{0}}

\global\long\def\bone{\boldsymbol{1}}

\global\long\def\balpha{\boldsymbol{\alpha}}

\global\long\def\bbeta{\boldsymbol{\beta}}

\global\long\def\bpi{\boldsymbol{\pi}}

\global\long\def\bphi{\boldsymbol{\phi}}

\global\long\def\bepsilon{\boldsymbol{\epsilon}}

\global\long\def\btheta{\boldsymbol{\theta}}

\global\long\def\bgamma{\boldsymbol{\gamma}}

\global\long\def\bPsi{\boldsymbol{\Psi}}

\global\long\def\bLambda{\boldsymbol{\Lambda}}

\global\long\def\bPhi{\boldsymbol{\Phi}}

\global\long\def\bmu{\boldsymbol{\mu}}

\global\long\def\tmu{\tilde{\mu}}

\global\long\def\btmu{\tilde{\bmu}}

\global\long\def\bamu{\bar{\mu}}

\global\long\def\bbmu{\bar{\bmu}}

\global\long\def\tlamda{\tilde{\lambda}}

\global\long\def\balamda{\bar{\lambda}}

\global\long\def\bSigma{\boldsymbol{\Sigma}}

\global\long\def\oo{\infty}

\global\long\def\bdelta{\boldsymbol{\delta}}

\global\long\def\boeta{\boldsymbol{\eta}}

\global\long\def\bsigma{\boldsymbol{\sigma}}

\global\long\def\given{\ |\ }

\global\long\def\goto{\rightarrow}

\global\long\def\asgoto{\stackrel{a.s.}{\longrightarrow}}

\global\long\def\pgoto{\stackrel{p}{\longrightarrow}}

\global\long\def\idp{\ \bot\negthickspace\negthickspace\bot\ }

\global\long\def\cdp{\idp}

\global\long\def\das{\triangleq}

\global\long\def\realset{\mathbb{R}}

\global\long\def\integerset{\mathbb{Z}}

\global\long\def\partialdev#1#2{\frac{\partial#1}{\partial#2}}

\global\long\def\partialdevdev#1#2{\frac{\partial^{2}#1}{\partial#2\partial#2^{\top}}}

\global\long\def\cpr#1#2{\Pr\left(#1\gv#2\right)}

\global\long\def\pr#1{\Pr\left( #1 \right)}

\global\long\def\transpose#1{#1^{\mathsf{T}}}

\global\long\def\invt#1{#1^{-1}}

\global\long\def\comb#1#2{\left({#1\atop #2}\right)}

\global\long\def\idmat{\mathbb{I}}

\global\long\def\id#1#2{\delta_{#1}^{(#2)}}

\global\long\def\argmax#1{\underset{_{#1}}{\text{argmax}} }

\global\long\def\argmin#1{\underset{_{#1}}{\text{argmin}\ } }

\global\long\def\SST{\mathbb{T}}

\global\long\def\SSTarg#1#2{\underset{_{#2}}{\SST}\left( #1 \right) }

\global\long\def\EXPV#1{\ < #1\  >}

\global\long\def\ESS{\mathbb{E}}

\global\long\def\ESSarg#1#2{\underset{_{#2}}{\ESS[}#1]}

\global\long\def\VAR{\text{Var}}

\global\long\def\VARarg#1#2{\underset{_{#2}}{\VAR}\left( #1 \right) }

\global\long\def\COV{\text{Cov}}

\global\long\def\vectorize#1{\mathbf{#1}}

\global\long\def\trace{\text{tr}}

\global\long\def\lcabra#1{\left|#1\right.}

\global\long\def\rcabra#1{\left.#1\right|}

\global\long\def\norm#1{\left\Vert #1\right\Vert }

\global\long\def\gaussden#1#2{\mathcal{N}\left(#1, #2 \right) }

\global\long\def\gausspdf#1#2#3{\mathcal{N}\left( #1 \lcabra{#2, #3}\right) }

\global\long\def\model{\mbox{NRBM}}

\title{Nonnegative Restricted Boltzmann Machines for Parts-based Representations
Discovery and \\Predictive Model Stabilization}

\author{Tu Dinh Nguyen, Truyen Tran, Dinh Phung, Svetha Venkatesh}
\maketitle
\begin{abstract}
The success of any machine learning system depends critically on effective
representations of data. In many cases, it is desirable that a representation
scheme uncovers the parts-based, additive nature of the data. Of current
representation learning schemes, restricted Boltzmann machines (RBMs)
have proved to be highly effective in unsupervised settings. However,
when it comes to parts-based discovery, RBMs do not usually produce
satisfactory results. We enhance such capacity of RBMs by introducing
nonnegativity into the model weights, resulting in a variant called
\emph{nonnegative restricted Boltzmann machine} (NRBM). The NRBM produces
not only controllable decomposition of data into interpretable parts
but also offers a way to estimate the intrinsic nonlinear dimensionality
of data, and helps to stabilize linear predictive models. We demonstrate
the capacity of our model on applications such as handwritten digit
recognition, face recognition, document classification and patient
readmission prognosis. The decomposition quality on images is comparable
with or better than what produced by the nonnegative matrix factorization
(NMF), and the thematic features uncovered from text are qualitatively
interpretable in a similar manner to that of the latent Dirichlet
allocation (LDA). The stability performance of feature selection on
medical data is better than RBM and competitive with NMF. The learned
features, when used for classification, are more discriminative than
those discovered by both NMF and LDA and comparable with those by
RBM.

\end{abstract}

\section{Introduction\label{sec:intro}}

Learning meaningful representations from data is often critical to
achieve high performance in machine learning tasks \cite{bengio_etal_pami13_representation}.
An attractive approach is to estimate representations that best explain
the data without the need for labels. One important class of such
methods is the restricted Boltzmann machine (RBM) \cite{smolensky_mit86_information,freund_haussler_techrep94_unsupervised},
an undirected probabilistic bipartite model in which a representational
hidden layer is connected with a visible data layer. The weights associated
with connections encode the strength of influence between hidden and
visible units. Each unit in the hidden layer acts as a binary feature
detector, and together, all the hidden units are linearly combined
through the connection weights to form a \emph{fully distributed representation}
of data \cite{hinton_ghahramani_biosci97_generative,bengio_etal_pami13_representation}.
This distributed representation is highly compact: for $\K$ units,
there are $2^{\K}-1$ non-empty configurations that can explain the
data.

\begin{figure}[h]
\noindent \centering{}\subfloat[RBM\label{fig:exp_ORL_filter_RBM}]{\noindent \centering{}\includegraphics[width=0.3\textwidth]{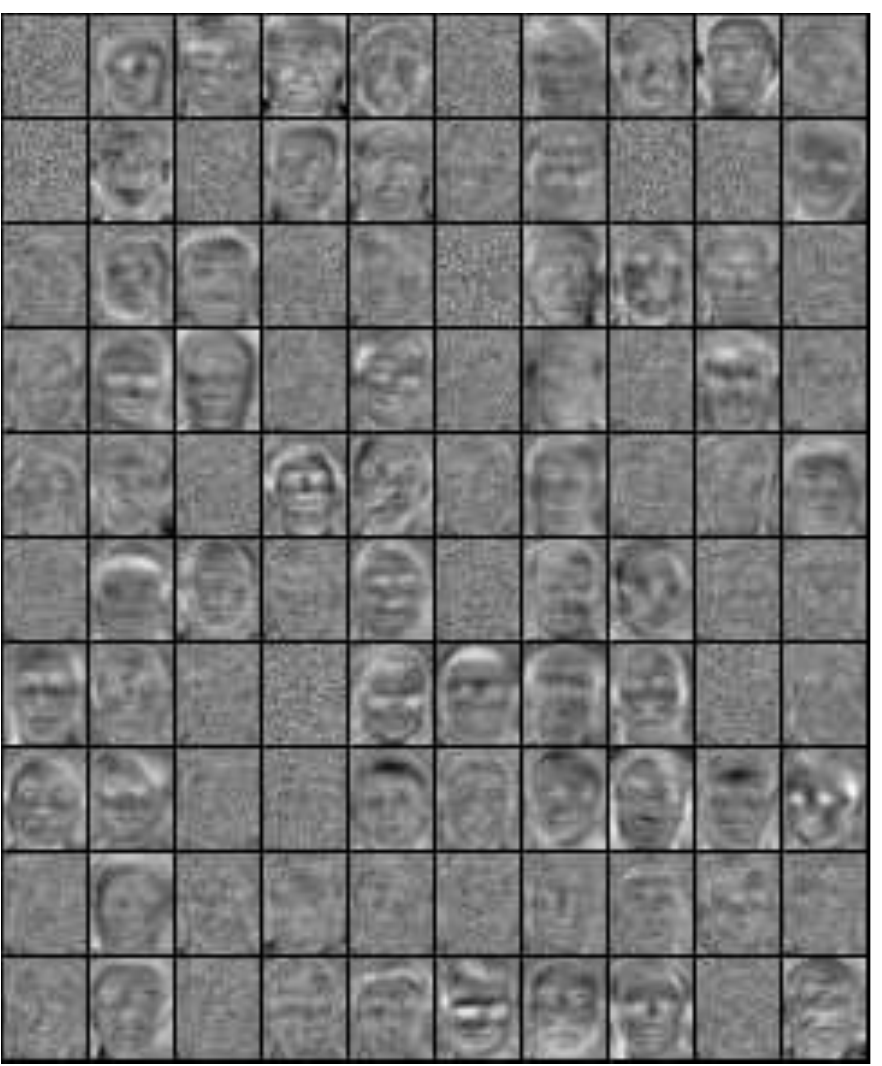}}\hspace{0.02\textwidth}\subfloat[NMF\label{fig:exp_ORL_filter_NMF}]{\noindent \centering{}\includegraphics[width=0.3\textwidth]{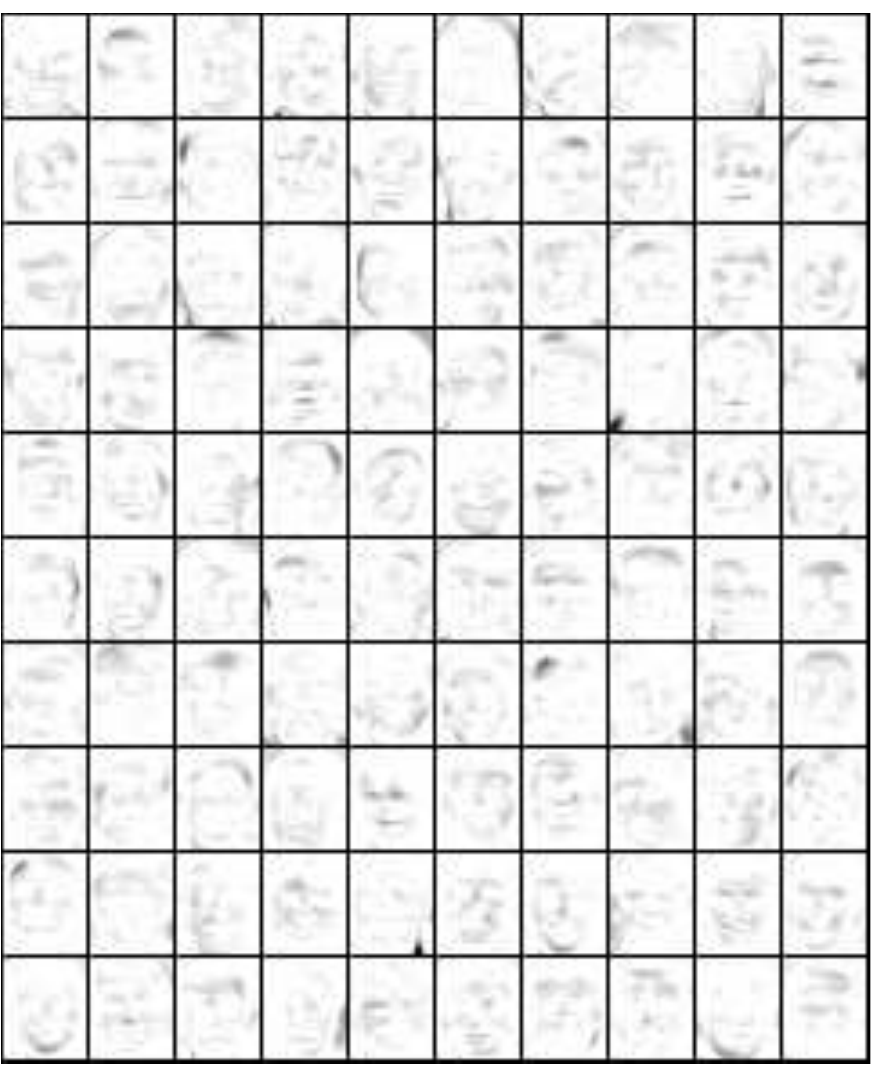}}\hspace{0.02\textwidth}\subfloat[$\protect\model$\label{fig:exp_ORL_filter_NRBM}]{\noindent \centering{}\includegraphics[width=0.3\textwidth]{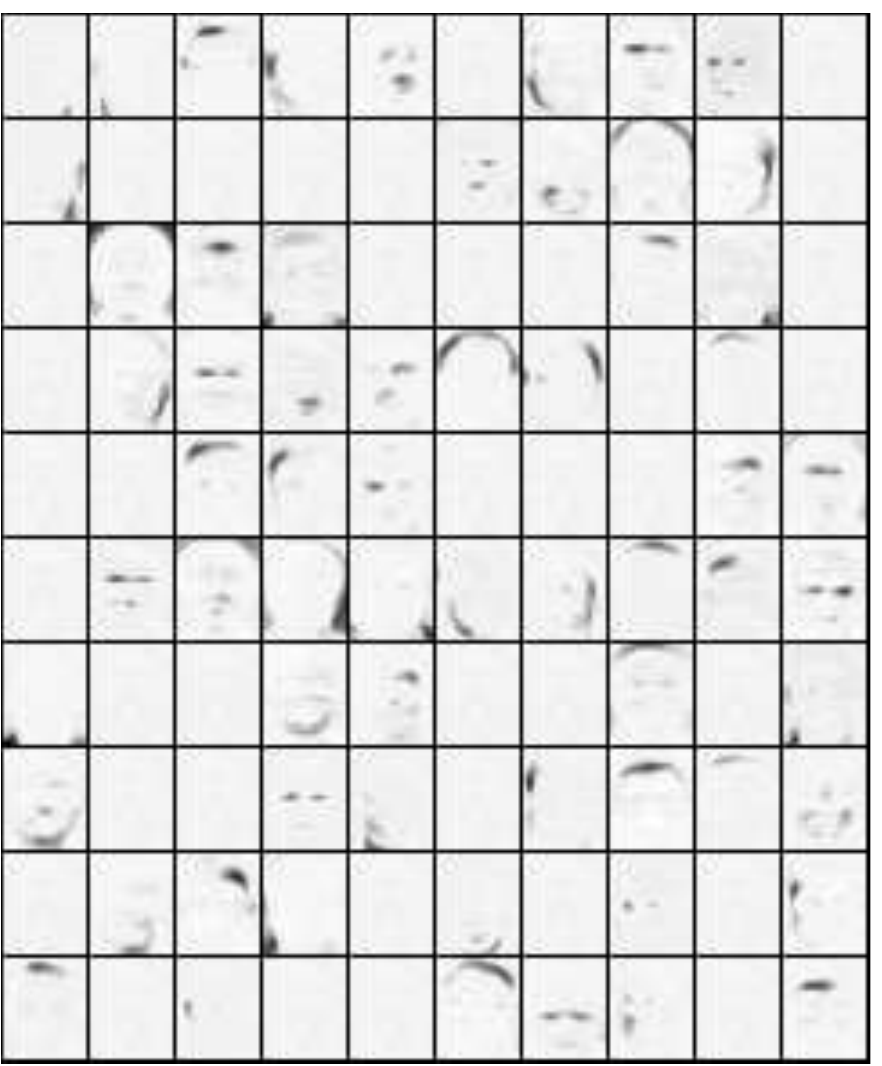}}\caption{Representations learned from the ORL face image database \cite{att_orl}
using the ordinary RBM, NMF and $\protect\model$ on the left, middle,
right, respectively. Darker pixels show larger weights.\label{fig:exp_ORL_filter}}
\end{figure}

However, the fully distributed representation learned by RBMs may
not interpretably disentangle the factors of variation since the learned
features are often global, that is, all the data units must play the
role in one particular feature. As a result, learned features do not
generally represent parts and components \cite{teh_etal_nips01_rbmrate}.
For example, the facial features learned by RBM depicted in Fig.~\ref{fig:exp_ORL_filter_RBM}
are generally global; it is hard to explain how a face is constructed
from these parts. Parts-based representations, on the other hand,
are perceptually intuitive. Using the same facial example, if individual
parts of the face (e.g. eyes, nose, mouth, forehead as in Fig.~\ref{fig:exp_ORL_filter_NRBM})
are discovered, it would be easy to construct the face from these
parts. In terms of modeling, detecting parts-based representation
also improves the object recognition performance \cite{agarwal_etal_pami04_learning}.

One of the best known techniques to achieve parts-based representation
is nonnegative matrix factorization (NMF) \cite{lee_etal_nature99_nmf}.
In NMF, the data matrix is approximately factorized into a basis matrix
and a coding matrix, where all the matrices are assumed to be nonnegative.
Each column of the basis matrix is a learned feature, which could
be sparse under appropriate regularization \cite{hoyer_jmlr04_nmf}.
The NMF, however, has a fundamental drawback: it does not generalize
to unseen data since there is no mechanism by which a new data point
can be generated from the learned model. Instead, new representations
must be learned from the expensive ``fold-in'' procedure. The RBM,
on the other hand, is a proper generative model -- once the model
has been learned, new samples can be drawn from the model distribution.
Moreover, due to the special bipartite structure, estimating representation
from data is efficient with a single matrix operation.

In this paper, we derive a novel method based on the RBM to discover
useful parts whilst retaining its discriminative capacity. As inspired
by the NMF, we propose to enforce nonnegativity in the connection
weight matrix of the RBM. Our method integrates a barrier function
into the objective function so that the learning is skewed towards
nonnegative weights. As the contribution of the visible units towards
a hidden unit is additive, there exists competition among visible
units to activate the hidden unit leading to a small portion of connections
surviving. In the same facial example, the method could achieve parts-based
representation of faces, which is, surprisingly, even better than
what learned by the standard NMF (cf. Fig.~\ref{fig:exp_ORL_filter}).
We term the resulting model the \emph{nonnegative restricted Boltzmann
machine} ($\model$).

In addition to parts-based representation, there are several benefits
with this nonnegativity constraint. First, in many cases, it is often
easier to make sense of addition of new latent factors (due to nonnegative
weights) than of subtraction (due to negative weights). For instance,
clinicians may be more comfortable with the notion that a risk factor
either contributes positively to a disease development or not at all
(e.g., the connections have zeros weights). Second, as weights can
be either positive or zero, the parameter space is highly constrained
leading to potential robustness. This can be helpful when there are
many more hidden units than those required to represent all factors
of variation: extra hidden units will automatically be declared ``dead''
if all connections to them cannot compete against others in explaining
the data. Lastly, combining two previous advantages, the $\model$
effectively gathers related important features, e.g., key risk factors
that explain the disease, into groups at the data layer and encourages
less hidden units to be useful at hidden layer. This helps to stabilize
linear predictive models by providing hidden representations for them
to perform on.

We demonstrate the effectiveness of the proposed model through comprehensive
evaluation on three real-world applications using five real datasets
of very different natures.
\begin{itemize}
\item The first application is parts-based discovery. Our primary targets
are to decompose images into interpretable parts (and receptive fields),
e.g., dots and strokes in handwritten digits (MNIST dataset \cite{lecun_etal_mnist}),
and facial components in faces (CBCL \cite{mit_cbcl} and ORL \cite{att_orl}
databases); and to discover plausible latent thematic features, which
are groups of semantically related words (TDT2 text corpus \cite{cai_etal_tkde05_document}).
\item The second application is feature extraction for classification. Here
we apply our models on MNIST, TDT2 and heart failure datasets. The
learned features are then fed into standard classifiers. The experiments
reveal that the classification performance is comparable with the
standard RBM, and competitive against NMF (on all images, text, and
medical records) and latent Dirichlet allocation (on text) \cite{blei_etal_jmlr03_lda}.
\item The last one is linear classifier stabilization. The goal is to enhance
feature and model stabilities in clinical prognosis. The experimental
results on heart failure patients show significant improvements of
stability scores over linear sparse model and the RBM, and better
prediction performances than those of RBM and NMF. This application
in healthcare analytics is the main extension of our previous model
introduced in \cite{tu_etal_acml13_nrbm}.
\end{itemize}
In short, our main contributions are: (i) the derivation of the nonnegative
restricted Boltzmann machine, a probabilistic machinery that has the
capacity of learning parts-based representations; (ii) a comprehensive
evaluation the capability of our method as a representational learning
tool on image, text and medical data, both qualitatively and quantitatively;
and (iii) a demonstration of stabilizing linear classifiers in clinical
prognosis.

The rest of the paper is structured as follows. Section~\ref{sec:nrbm}
presents the derivation and properties of our nonnegative RBM, followed
by its applications in linear predictive model stabilization in Section~\ref{sec:nrbm_stab}.
We then report our experimental results in Section~\ref{sec:result}.
Section~\ref{sec:related_work} provides a discussion of the related
literature. Finally, Section~\ref{sec:conclusion} concludes the
paper.

\section{Preliminaries}

\begin{figure}[h]
\noindent \centering{}\includegraphics[width=0.5\textwidth]{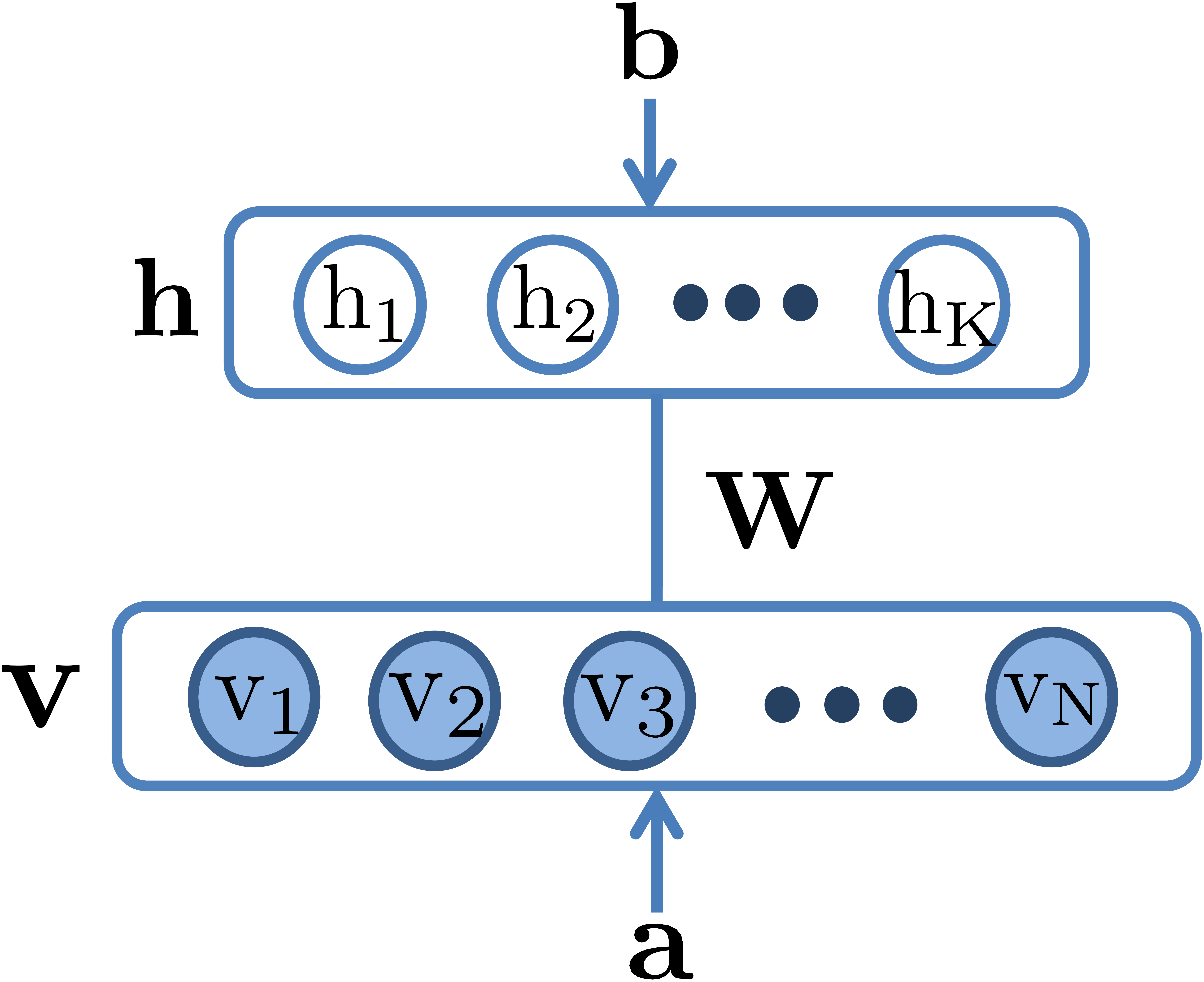}\caption{Graphical illustration of a RBM that models the joint distribution
of $\protect\N$ visible units and $\protect\K$ hidden units. The
connections are undirected and the shaded nodes are observed.\label{fig:RBM_graphical_model}}
\end{figure}

We first describe the restricted Boltzmann machine (RBM) for unsupervised
learning representation. An RBM \cite{smolensky_mit86_information,freund_haussler_techrep94_unsupervised,hinton_neucom02_cd}
is a bipartite undirected graphical model in which the bottom layer
contains observed variables called visible units and the top layer
consists of latent \emph{representational variables}, known as hidden
units. Two layers are fully connected but there is no connection within
layers. The visible units can model the data while the hidden units
can capture the latent factors not presented in the observations.
The hidden units are linearly combined through connection weights.
A graphical illustration of RBM is presented in Fig.~\ref{fig:RBM_graphical_model}.
As a matter of convention in the literature of RBM, we shall use the
term \textquotedblleft unit\textquotedblright{} and \textquotedblleft random
variable\textquotedblright{} interchangeably.

\subsection{Model representation}

Let $\bv$ denote the set of visible variables: $\bv=\left[\v_{1},\v_{2},...,\v_{\N}\right]^{\top}\in\left\{ 0,1\right\} ^{\N}$
and $\bh$ indicate the set of hidden ones: $\bh=\left[\h_{1},\h_{2},...,\h_{\K}\right]^{\top}\in\left\{ 0,1\right\} ^{\K}$.
The RBM defines an energy function of a joint configuration $\left(\bv,\bh\right)$
as:
\begin{align}
E\left(\bv,\bh;\psi\right) & =-\left(\ba^{\top}\bv+\bb^{\top}\bh+\bv^{\top}\bW\bh\right)\label{eq:RBM_energy}
\end{align}

\noindent where $\psi=\left\{ \ba\mathbf{,}\bb,\bW\right\} $ is the
set of parameters. $\ba=\left[\a_{n}\right]_{\N}\in\realset^{\N},\bb=\left[\b_{k}\right]_{\K}\in\realset^{\K}$
are the biases of hidden and visible units respectively, and $\bW=\left[\w_{nk}\right]_{\N\times\K}\in\realset^{\N\times\K}$
represents the weights connecting the hidden and visible units. The
model assigns a Boltzmann distribution (also known as Gibbs distribution)
to the joint configuration as:
\begin{align}
p\left(\bv,\bh;\psi\right) & =\frac{1}{\mathcal{Z}\left(\psi\right)}e^{-E\left(\bv,\bh;\psi\right)}\label{eq:RBM_joint_prob}
\end{align}

\noindent where $\mathcal{Z}\left(\psi\right)$ is the normalization
constant, computed by summing over all possible states pairs of visible
and hidden units:
\begin{align*}
\partf\left(\psi\right) & =\sum_{\bv,\bh}e^{-E\left(\bv,\bh;\psi\right)}
\end{align*}

Since the network has no intra-layer connections, the Markov blanket
of each unit contains only the units of the other layer. Units in
one layer become conditionally independent given the other layer.
Thus the conditional distributions over visible and hidden units are
nicely factorized as:
\begin{align}
p\left(\bv\mid\bh;\psi\right) & =\prod_{n=1}^{\N}p\left(\v_{n}\mid\bh;\psi\right)\label{eq:RBM_visfac}\\
p\left(\bh\mid\bv;\psi\right) & =\prod_{k=1}^{\K}p\left(\h_{k}\mid\bv;\psi\right)\label{eq:RBM_hidfac}
\end{align}
wherein the conditional probabilities of single units being active
are:
\begin{align}
p\left(\v_{n}=1\gv\bh;\psi\right) & =\sig\left(\a_{n}+\bw_{n\bigcdot}\bh\right)\nonumber \\
p\left(\h_{k}=1\gv\bv;\psi\right) & =\sig\left(\b_{k}+\bv^{\top}\bw_{\bigcdot k}\right)\label{eq:RBM_hidprob}
\end{align}
with the logistic sigmoid function $\sig\left(x\right)=\left[1+e^{-x}\right]^{-1}$.

\subsection{Parameter learning}

The parameter learning of RBM is performed by maximizing the following
data log-likelihood:
\begin{align*}
\likelihood\left(\bv;\psi\right) & =\log p\left(\bv;\psi\right)=\log\sum_{\bh}p\left(\bv,\bh;\psi\right)
\end{align*}
The parameters are updated using stochastic gradient ascent as follows:
\begin{align*}
\w_{nk} & \leftarrow\w_{nk}+\eta\left(\expect_{\tilde{p}}\left[\v_{n}\h_{k}\right]-\expect_{p}\left[\v_{n}\h_{k}\right]\right)
\end{align*}
wherein $\eta>0$ is the learning rate, $\tilde{p}\left(\bv,\bh;\psi\right)=\tilde{p}\left(\bv;\psi\right)p\left(\bh\gv\bv;\psi\right)$
is the data distribution with $\tilde{p}\left(\bv;\psi\right)$ representing
the empirical distribution, $p\left(\bv,\bh;\psi\right)$ is the model
distribution defined in Eq.~(\ref{eq:RBM_joint_prob}). Whilst the
data expectation $\expect_{\tilde{p}}\left[\bigcdot\right]$ can be
computed efficiently, the model expectation $\expect_{p}\left[\bigcdot\right]$
is intractable. To overcome this shortcoming, we use a truncated MCMC-based
method known as contrastive divergence (CD) \cite{hinton_neucom02_cd}
to approximate the model expectation. This approximation approach
is efficient since the factorizations in Eqs.~(\ref{eq:RBM_visfac},\ref{eq:RBM_hidfac})
allow fast layer-wise sampling.

\subsection{Representation learning}

Once the model is fully specified, the new representation of an input
data can be achieved by computing the posterior vector $\hat{\bh}=\left[\hh_{1},\hh_{2},...,\hh_{\K}\right]$,
where $\hh_{\K}$ is shorthand for $p\left(\h_{k}=1\mid\bv;\psi\right)$
as given in Eq.~(\ref{eq:RBM_hidprob}).

\section{Nonnegative restricted Boltzmann machine\label{sec:nrbm}}

We now present our main contribution -- the nonnegative restricted
Boltzmann machine ($\model$) that integrates nonnegativity into the
connection weights of the model. We then present the capability of
$\model$ to estimate the intrinsic dimensionality of the data, and
to stabilize linear predictive models.

\subsection{Deriving parts-based representation\label{sub:nrbm_derive}}

The derivation of parts-based representation starts from the connection
weights of the standard RBM. In the RBM, two layers are connected
using a weight matrix $\bW=\left[\w_{nk}\right]_{\N\times\K}$ in
which $\w_{nk}$ is the association strength between the hidden unit
$k$ and the visible unit $n$. The column vector $\bw_{\bigcdot k}$
is the learned filter of the hidden unit $k$. Parts-based representations
imply that this column vector must be sparse, e.g. only a small portion
of entries is non-zeros. Recall that the activation of this hidden
unit, also known as the firing rate in neural network language, is
the probability: $p\left(\h_{k}=1\mid\bv;\psi\right)=\sigma\left(\b_{k}+\sum_{n=1}^{\N}\w_{nk}\v_{n}\right)$.
The positive connection weights tend to activate the associated hidden
units whilst the negative turn the units off. In addition, the positive
weights add up to representations whereas the negative subtract. Thus
it is hard to determine which factors primarily contribute to the
learned filter.

\begin{figure}[h]
\noindent \centering{}\includegraphics[width=0.6\textwidth]{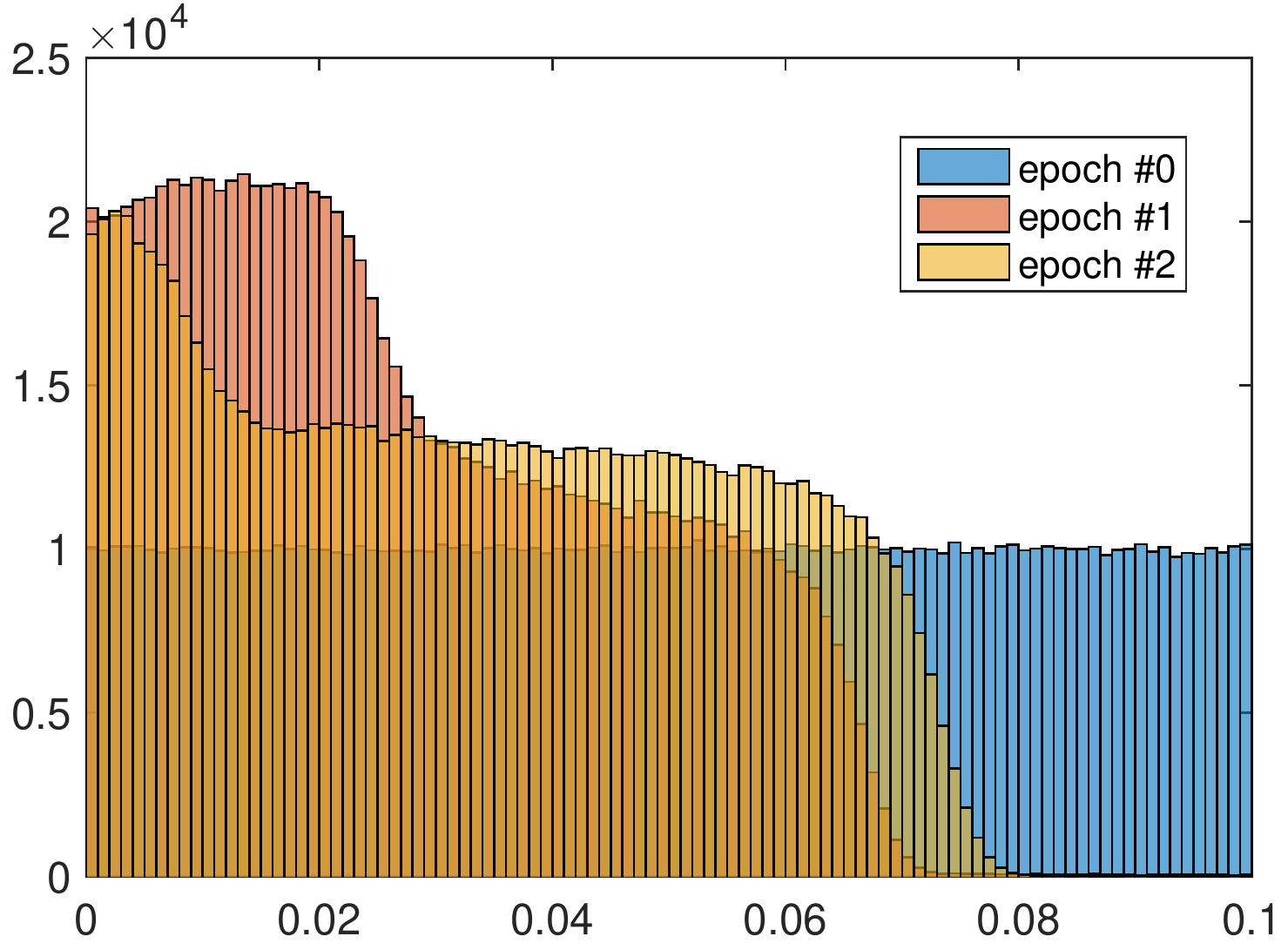}\caption{The histograms of NRBM's weights at the initialization (epoch $\#0$)
and learned at epoch $\#1$ and $\#2$.\label{fig:nrbm_whist}}
\end{figure}

Due to asymmetric parameter initialization, the learning process tends
to increase some associations more than others. Under nonnegative
weights, i.e. $\w_{nk}\ge0$, the hidden and visible units tend to
be more co-active. One can expect that for a given activation probability
$\hh_{k}$ and bias $\b_{k}$, such increase must cause other associations
to degrade, since $\v_{n}\ge0$. As the lower bounds of weights are
now zeros, there is a natural tendency for many weights to be driven
to zeros as the learning progresses. For an illustration, Fig.~\ref{fig:nrbm_whist}
shows the histograms of model's weights at initialization (epoch $\#0$)
and learned at epoch $\#1$ and $\#2$. It can be seen that, the number
of weights close to zeros increases after each learning epoch.

Recall that learning in the standard RBM is usually based on maximizing
the data likelihood with respect to the parameter $\psi$ as:
\begin{align}
\mathcal{L}\left(\bv;\psi\right) & =p\left(\bv;\psi\right)=\sum_{\bh}p\left(\bv,\bh;\psi\right)\label{eq:RBM_likelihood}
\end{align}
To encourage nonnegativity in $\bW$, we integrate a nonnegative constraint
under quadratic barrier function \cite{nocedal_wright_springer99_numerical}
into the learning of the standard RBM. The objective function now
turns into the following regularized log-likelihood:
\begin{align}
\likelihood_{\textrm{reg}} & =\log\mathcal{L}\left(\bv;\psi\right)-\frac{\alpha}{2}\sum_{n=1}^{\N}\sum_{k=1}^{\K}f\left(\w_{nk}\right)\label{eq:NRBM_reg_loglikelihood}
\end{align}
where:
\begin{align*}
f\left(\x\right) & =\begin{cases}
\x^{2} & \x<0\\
0 & \x\geq0
\end{cases}
\end{align*}
and $\alpha\geq0$. The penalty of sum squared of negative weights
regularizes the optimization to encourage nonnegativity in $\bW$.
The learning procedure now performs minimizing the penalty term and
maximizing the data log-likelihood $\log\likelihood\left(\psi;\bv\right)$
simultaneously to achieve the optimal solution for the regularized
function. The degree of nonnegativity is controlled by the hyperparameter
$\alpha$. If we set $\alpha$ to zero, the barrier function will
vanish and the model will revert to the ordinary RBM without nonnegative
constraint. The larger $\alpha$ is, the tighter the barrier restriction
becomes.

Finally the parameter update rule reads:
\begin{equation}
\w_{nk}\leftarrow\w_{nk}+\eta\left(\left\langle \v_{n}\h_{k}\right\rangle _{\tilde{p}}-\left\langle \v_{n}\h_{k}\right\rangle _{p}-\alpha\left\lceil \w_{nk}\right\rceil ^{-}\right)\label{eq:NRBM_update_param}
\end{equation}
where $\eta>0$ is the learning rate, $\tilde{p}\left(\bv,\bh;\psi\right)=\tilde{p}\left(\bv;\psi\right)p\left(\bh\mid\bv;\psi\right)$
is the data distribution with $\tilde{p}\left(\bv;\psi\right)$ representing
the empirical distribution, $p\left(\bv,\bh;\psi\right)$ is the model
distribution defined in Eq.~(\ref{eq:RBM_joint_prob}) and $\left\lceil \w_{nk}\right\rceil ^{-}$
denotes the negative part of the weight. Following the learning of
standard RBMs \cite{hinton_neucom02_cd}, $\tilde{p}\left(\bv;\psi\right)$
is computed from data observations and $p\left(\bv,\bh;\psi\right)$
can be efficiently approximated by running multiple Markov short chains
starting from observed data at each update.

\subsection{Estimating intrinsic dimensionality\label{sub:nrbm_intrinsic_dim}}

The fundamental question of the RBM is how many hidden units are enough
to model the data. Currently, no easy methods determine the appropriate
number of hidden units needed for a particular problem. One often
uses a plentiful number of units, leading to some units are redundant.
Note that the unused hidden units problem of RBM is not uncommon which
has been studied in \cite{berglund_etal_iconip13_measuring}. If the
number of hidden units is lower than the number of data features,
the RBM plays a role of a dimensionality reduction tool. This suggests
one prominent method -- principal component analysis (PCA) which captures
the data variance using the principal components. The amount of variance,
however, can help the method specify the number of necessary components.
The nonnegativity constraint in the $\model$, interestingly, leads
to a similar capacity by examining the ``dead'' hidden units.

To see how, recall that the hidden and visible units are co-active
via the connection weights:
\begin{align}
p\left(\v_{n}=1\mid\bh\right) & =\sigma\left(\a_{n}+\sum_{k=1}^{\K}\w_{nk}\h_{k}\right)\label{eq:RBM_visprob}
\end{align}
Since this probability, in general, is constrained by the data variations,
the hidden units must ``compete'' with each other to explain the
data. This is because the contribution towards the explanation is
nonnegative, thus an increase on the power of one unit must be at
the cost of others. If $\K^{*}<\K$ hidden units are intrinsically
enough to account for all variations in the data, then one can expect
that either the other $\K-\K^{*}$ hidden units are always deactivated
(e.g. with very large negative biases) or their connection weights
are almost zeros since $\w_{nk}\ge0$. In either cases, the hidden
units become permanently inoperative in data generation. Thus by examining
the dead units, we may be able to uncover the intrinsic dimensionality
of the data variations.

\section{Stabilizing Linear Predictive Models\label{sec:nrbm_stab}}

This section presents an application of $\model$ in stabilizing linear
predictive models. Model stability is often overlooked in prediction
models which pay more attention to predictive performances. However,
model stability is as important as the prognosis in domains where
model parameters are interpreted by humans and are subject to external
validation. Particularly, in the medical domain, the model stability
increases reliability, interpretability, generalization (or transferability)
and reproducibility \cite{awada_etal_iri12_review,khoshgoftaar_etal_iri13_survey}.
For example, it is desirable to have a reliable model that can discover
a stable set of risk factors to interpret the causes of the disease.
The generalization of the method is the capability to transfer knowledge
from one disease to another. This also helps to improve clinical adoption.
Finally, stability enables researchers to reproduce the results easily.

A popular modern method to learn predictive risk factors from a high-dimensional
data is to employ $\ell_{1}$-penalty, following the success of lasso
in linear regression. Lasso is a sparse linear predictive model which
promotes feature shrinkage and selection \cite{tibshirani_jrss96_regression}.
More formally, let $\bX=\left[\x_{mn}\right]_{\M\times\N}\in\realset^{\M\times\N}$
denote the data matrix consisting of $\M$ data points with $\N$
features and $\by=\left[\y_{1},\y_{2},...,\y_{\M}\right]^{\top}\in\left\{ 0,1\right\} ^{\M}$
denote the labels. Consider a predictive distribution:
\begin{align}
p\left(\y\gv\bx;\bw\right) & =p\left(\y\gv\sum_{n=1}^{\N}\w_{n}\x_{n}\right)\label{eq:NRBM_modelstab_pdf}
\end{align}
Lasso-like learning optimizes a $\ell_{1}$-penalized log-likelihood:
\begin{align}
\likelihood_{\textrm{lasso}} & =\frac{1}{\M}\sum_{m=1}^{\M}\log p\left(\y_{m}\mid\bx_{m\bigcdot};\bw\right)-\beta\left\Vert \bw\right\Vert _{1}\label{eq:NRBM_modelstab_objfunc}
\end{align}
in which $\bw=\left[\w_{n}\right]_{\N\times1}$ is the parameter vector
and $\beta>0$ is the regularization hyperparameter. The regularization
induces the sparsity of weight vector $\bw$.

The lasso, however, is susceptible to data variations (e.g. resampling
by bootstrapping, slight perturbation), resulting in loss of stability
\cite{austin_tu_2004_automated,xu_etal_pami12_sparse}. The method
often chooses one of two highly correlated features, resulting in
only a 0.5 chance for strongly predictive feature pairs. To overcome
this problem, our solution is to provide the lasso with lower-dimensional
data representations whose features are expected to be less correlated.
Here we introduce a two-stage framework which is a pipeline with the
$\model$ followed by the lasso. The first stage is to learn the connection
weights $\bW=\left[\w_{nk}\right]_{\N\times\K}$ of $\model$. Then
the machine can map the data onto new representations, i.e. the hidden
posteriors: $\bX=\left[\x_{mn}\right]_{\M\times\N}\mapsto\bhX=\left[\hx_{mk}\right]_{\M\times\K}$
(cf. Eq.~(\ref{eq:RBM_hidprob})). The representations are in $\K$-dimensional
space and used as the new data for the lasso. At the second stage,
the shrinkage method learns $\K$ weights for $\K$ features (i.e.
hidden units) to predict the label. Suppose we obtain weight vector
$\bhw=\left[\hw_{1},\hw_{2},...,\hw_{\K}\right]^{\top}$, we have:
\begin{align*}
p\left(\y\gv\bhx;\bhw\right) & =p\left(\y\gv\sum_{k=1}^{\K}\hw_{k}\hx_{k}\right)\\
 & =p\left(\y\gv\sum_{k=1}^{\K}\hw_{k}\sum_{n=1}^{\N}\x_{n}\w_{nk}\right)\\
 & =p\left(\y\gv\sum_{n=1}^{\N}\x_{n}\sum_{k=1}^{\K}\hw_{k}\w_{nk}\right)
\end{align*}

It can be seen that we can sum over the multiplications of the weights
of two methods to obtain a new weight vector $\bbarw=\left[\barw_{1},\barw_{2},...,\barw_{\N}\right]^{\top}$
as below:
\begin{align}
\barw_{n} & =\sum_{k=1}^{\K}\hw_{k}\w_{nk}\label{eq:NRBM_modelstab_conjugated_w}
\end{align}
These weights connect original features to the label as the weights
of lasso in Eq.~(\ref{eq:NRBM_modelstab_pdf}). Thus they can be
used to assess the stability of our proposed framework.

\section{Experiments\label{sec:result}}

In this section, we quantitatively evaluate the capacity of our proposed
model -- nonnegative RBM on three applications:
\begin{itemize}
\item \emph{Parts-based discovery}: unsupervised decomposing images into
parts, discovering semantic features from texts, and grouping clinically
relevant features;
\item \emph{Feature extraction for classification}: discovering discriminative
features that help supervised classification; and
\item \emph{Linear predictive model stabilization}: stabilizing feature
selection towards rehospitalization prognosis.
\end{itemize}
We use five real-world datasets in total: three for images, one for
text and one for medical data. Three popular image datasets are: one
for handwritten digits -- MNIST \cite{lecun_etal_mnist} and two for
faces -- CBCL \cite{mit_cbcl} and ORL \cite{att_orl}. For these
datasets, our primary target is to decompose images into interpretable
parts (and receptive fields), e.g. dots and strokes in handwritten
digits, and facial components in faces. The text corpus is $30$ categories
subset of the TDT2 corpus\footnote{NIST Topic Detection and Tracking corpus is at \href{http://www.nist.gov/itl/}{http://www.nist.gov/itl/}.}.
The goal is to discover plausible latent thematic features, which
are groups of semantically related words. Lastly, the medical data
is a collection of heart failure patients provided by Barwon Health,
a regional health service provider in Victoria, Australia. We aim
to investigate the feature stability during the rehospitalization
prediction of patients. For prognosis, we use logistic regression
model, i.e, $p\left(\y\gv\bx;\bw\right)$ in Eq.~(\ref{eq:NRBM_modelstab_pdf})
is now the probability mass function of a Bernoulli distribution,
for 6-month readmission after heart failure.

\paragraph{Image datasets}

The MNIST dataset consists of $60,000$ training and $10,000$ testing
$28\times28$ images, each of which contains a handwritten digit.
The CBCL database contains facial and non-facial images wherein our
interest is only on $2,429$ facial images in the training set. The
images are histogram equalized, cropped and rescaled to a standard
form of $19\times19$. Moreover, the human face in each image is also
well-aligned. By contrast, the facial images of an individual subject
in ORL dataset are captured under a variety of illumination, facial
expressions (e.g. opened or closed eyes, smiling or not) and details
(e.g. glasses, beard). There are $10$ different images for each of
$40$ distinct people. Totally, the data consists of $400$ images
with the same size of $92\times112$ pixels. Images of the three datasets
are all in the grayscale whose pixel values are then normalized into
the range $[0,1]$. Since the image pixels are not exactly binary
data, following the previous work \cite{hinton_salakhutdinov_sci06_reducing},
we treat the normalized intensity as empirical probabilities on which
the $\model$ is naturally applied. As the empirical expectation $\left\langle \v_{n}\h_{k}\right\rangle _{\tilde{p}}$
in Eq.~(\ref{eq:NRBM_update_param}) requires the probability $p\left(\bv;\psi\right)$,
the normalized intensity is a good approximation.

\paragraph{Text dataset}

The TDT2 corpus is collected during the first half of $1998$ from
six news sources: two newswires (APW, NYT), two radio programs (VOA,
PRI), and two television programs (CNN, ABC). It contains $11,201$
on-topic documents arranged into $96$ semantic categories. Following
the preprocessing in \cite{cai_etal_tkde05_document}, we remove all
multiple category documents and keep the largest $30$ categories.
This retains $9,394$ documents and $36,771$ unique words in total.
We further the preprocessing of data by removing common stopwords.
Only $1,000$ most frequent words are then kept and one blank document
are removed. For $\model$, word presence is used rather than their
counts.

\paragraph{Heart failure data}

The data is collected from the Barwon Health which has been serving
more than $350,000$ residents. For each time of hospitalization,
patient information is recorded into a database using the MySQL server
of the hospital. Each record contains patient admissions and emergency
department (ED) attendances which form an electronic medical record
(EMR). Generally, each EMR consists of demographic information (e.g.
age, gender and postcode) and time-stamped events (e.g. hospitalizations,
ED visits, clinical tests, diagnoses, pathologies, medications and
treatments). Specifically, it includes international classification
of disease 10 (ICD-10) scheme \cite{WHO_ICD-10}, Australian invention
coding (ACHI) scheme \cite{ACHI7}, diagnosis-related group (DRG)
codes, detailed procedures and discharge medications for each admission
and ED visit. Ethics approval was obtained from the hospital and research
ethics committee at Barwon Health (number 12/83) and Deakin University.

For our study, we collect the retrospective data of heart failure
patients from the hospital's database. The resulting cohort contains
$1,405$ unique patients with $1,885$ admissions between January
2007 and December 2011. We identify patients as heart failure if they
had at least one ICD-10 diagnosis code I50 at any admission. Patients
of all age groups are included whilst inpatient deaths are excluded
from our cohort. Among these patients, $49.3\%$ are male and the
medium age is $81.5$ at the time of admission. We focus our study
on emergency attendances and unplanned admissions of patients. The
readmission of patients is defined as an admission within the horizons
of 1, 6 and 12 months after the prior discharge date. After retrieving
the data, we follow the \emph{one-sided convolutional filter bank}
method introduced in \cite{tran_etal_kdd13_integrated} to extract
features which are then normalized into the range $\left[0,1\right]$.

To speed up the training phase, we divide training samples into ``mini-batches''
of $B=100$ samples. Hidden, visible and visible-hidden learning rates
are fixed to $0.1$. Visible biases are initialized so that the marginal
distribution, when there are no hidden units, matches the empirical
distribution. Hidden biases are first set to some reasonable negative
values to offset the positive activating contribution from the visible
units. Mapping parameters are randomly drawn from positive values
in $[0,0.01]$. Parameters are then updated after every mini-batch.
Learning is terminated after $100$ epochs. The regularization hyperparameter
$\alpha$ in Eq.~(\ref{eq:NRBM_reg_loglikelihood}) is empirically
tuned so that the data decomposition is both meaningful (e.g. by examining
visually, or by computing the parts similarity) and accurate (e.g.
by examining the reconstruction quality). The hyperparameter $\beta$
in Eq.~(\ref{eq:NRBM_modelstab_objfunc}) is set to $0.001$, which
is to achieve the best prediction score.

\subsection{Part-based discovery}

\subsubsection{Decomposing images into parts-based representations\label{sub:exp_parts_decom}}

\begin{figure}
\noindent \centering{}\subfloat[RBM\label{fig:exp_mnist_filter_rbm}]{\noindent \begin{centering}
\includegraphics[width=0.4\textwidth]{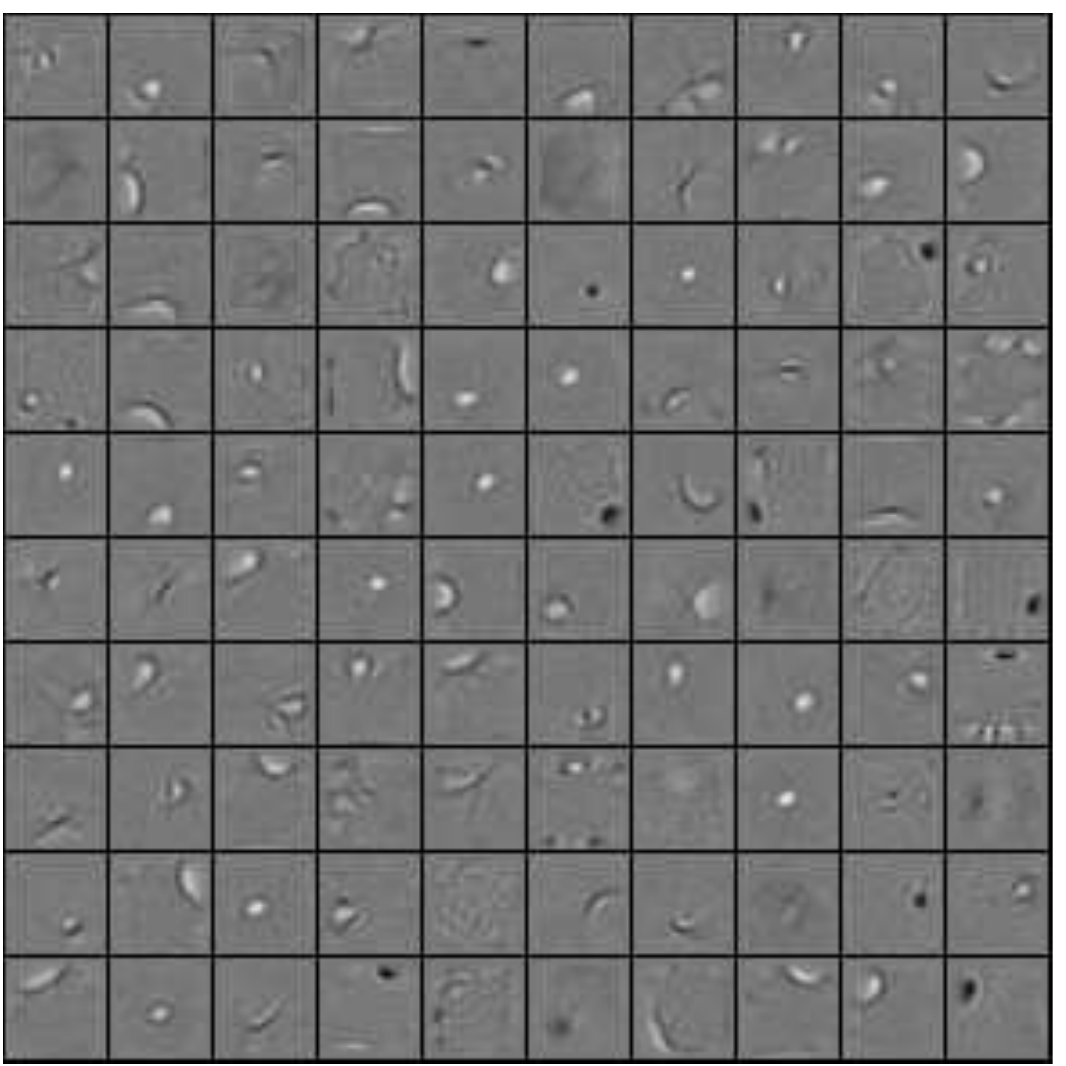}
\par\end{centering}

}\hspace{0.08\textwidth}\subfloat[$\protect\model$\label{fig:exp_mnist_filter_nrbm}]{\noindent \centering{}\includegraphics[width=0.4\textwidth]{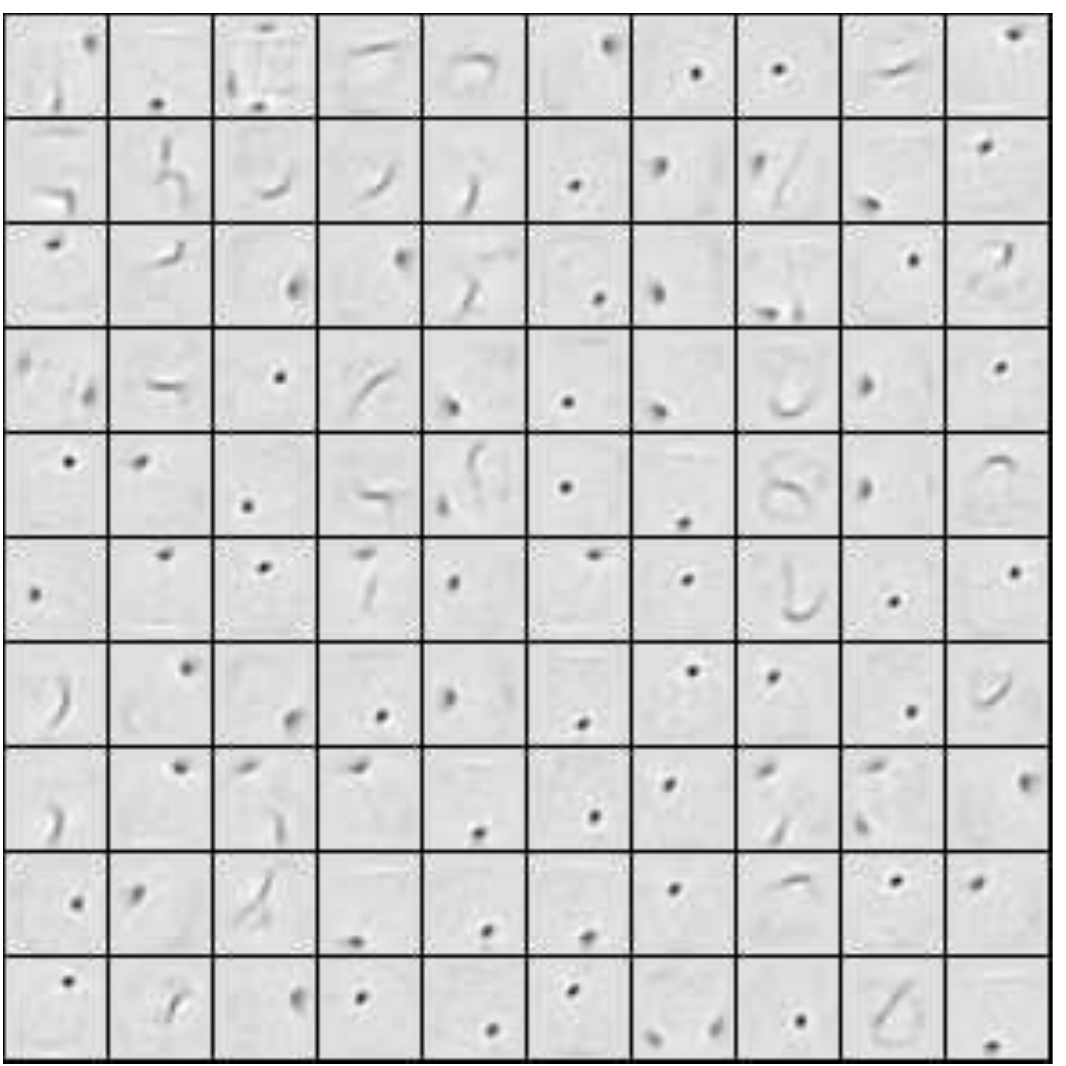}}\caption{Receptive fields learned from the MNIST handwritten digits database
using the RBM on Fig.~\ref{fig:exp_mnist_filter_rbm} and $\protect\model$
on Fig.~\ref{fig:exp_mnist_filter_nrbm}. Darker pixels illustrate
larger weights.\label{fig:exp_mnist_filter}}
\end{figure}

We now show that the nonnegative constraints enable the $\model$
to produce meaningful parts-based receptive fields. Fig.~\ref{fig:exp_mnist_filter}
depicts the $100$ filters learned from the MNIST images. It can be
seen that basic structures of handwritten digits such as strokes and
dots are discovered by both RBM and $\model$. However, the features
that $\model$ learns on Fig.~\ref{fig:exp_mnist_filter_nrbm} are
simpler whilst the ones learned by RBM on Fig.~\ref{fig:exp_mnist_filter_rbm}
are more difficult to interpret.

\begin{figure}
\noindent \centering{}\subfloat[RBM\label{fig:exp_cbcl_filter_rbm}]{\noindent \centering{}\includegraphics[width=0.3\textwidth]{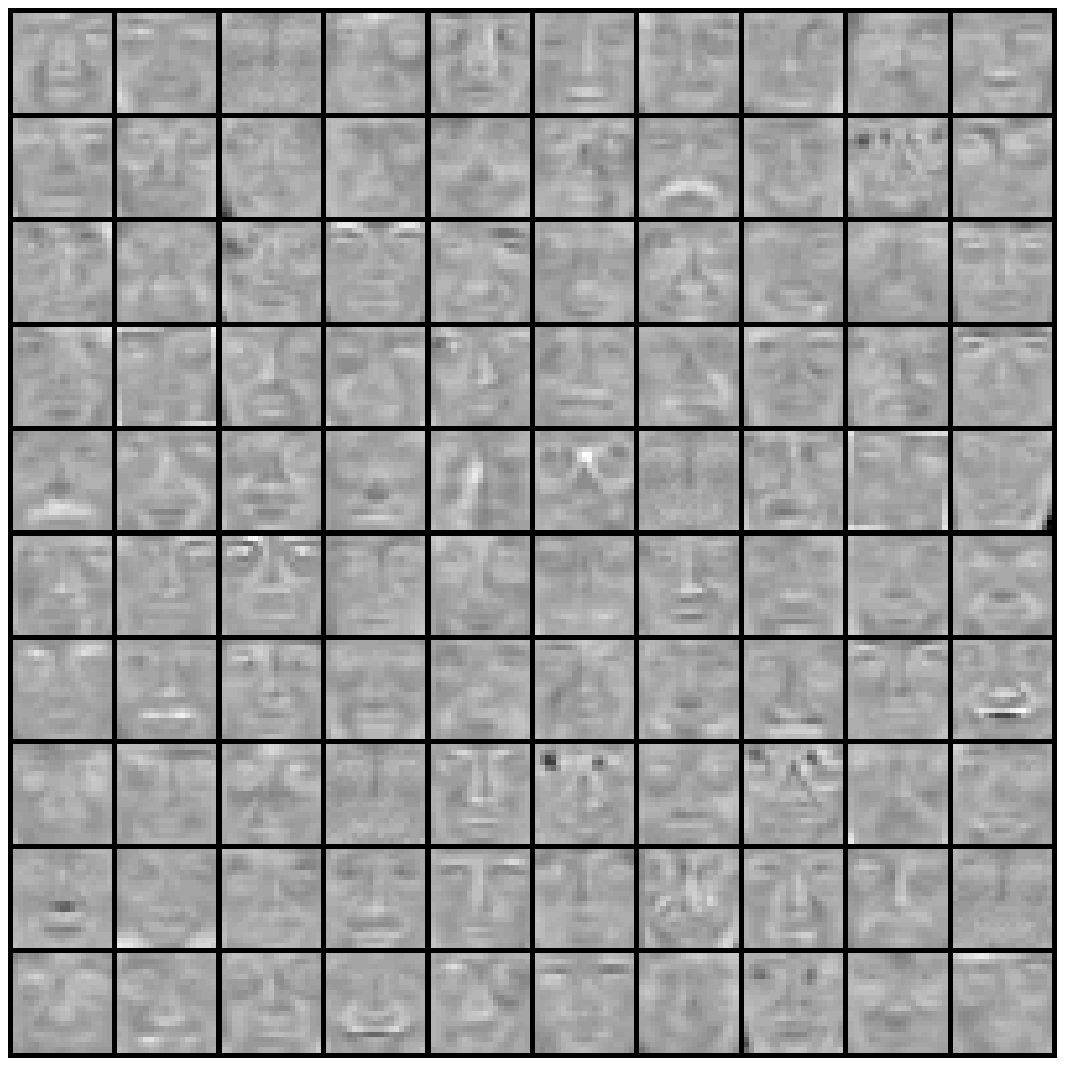}}\hspace{0.02\textwidth}\subfloat[NMF\label{fig:exp_cbcl_filter_nmf}]{\noindent \centering{}\includegraphics[width=0.3\textwidth]{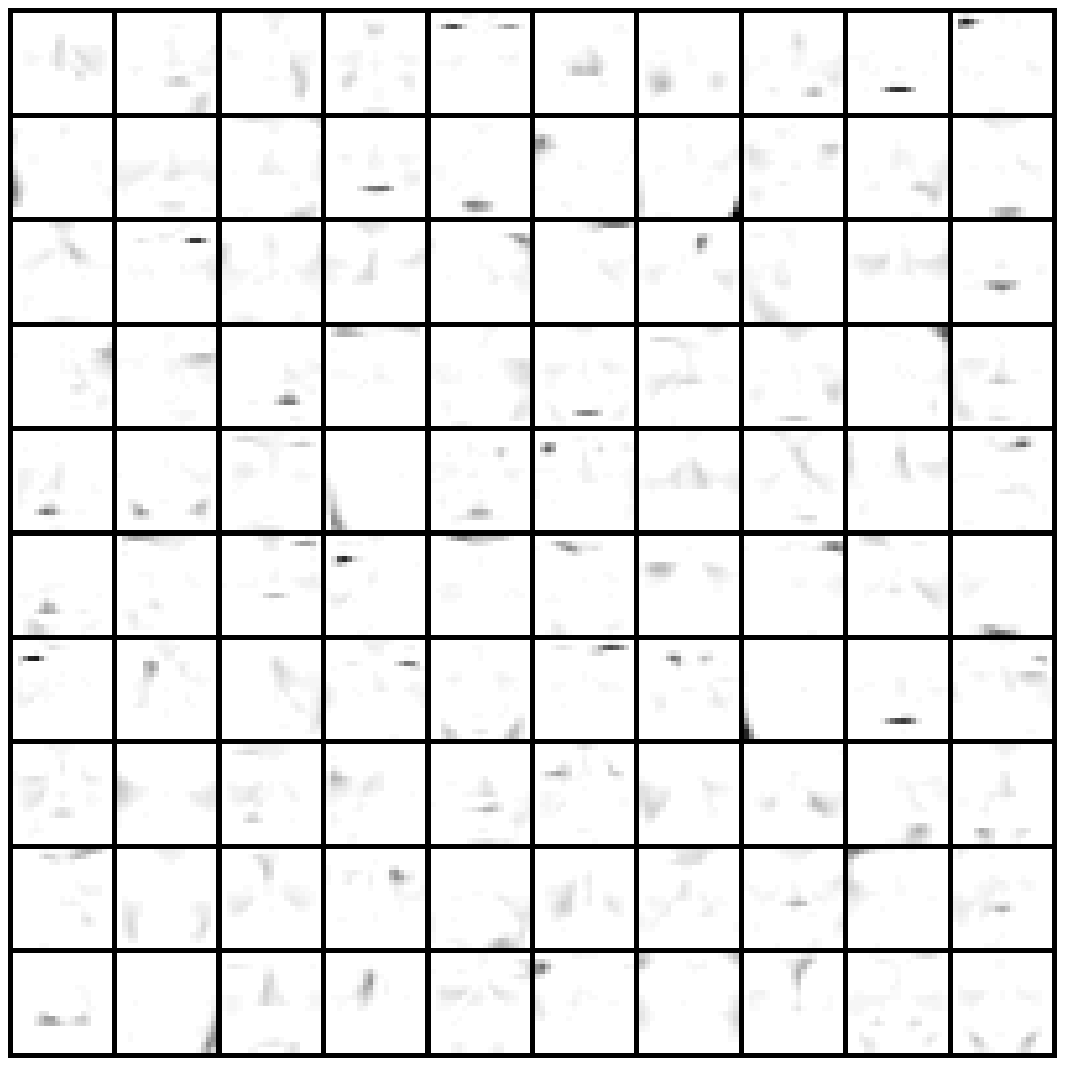}}\hspace{0.02\textwidth}\subfloat[$\protect\model$\label{fig:exp_cbcl_filter_nrbm}]{\noindent \centering{}\includegraphics[width=0.3\textwidth]{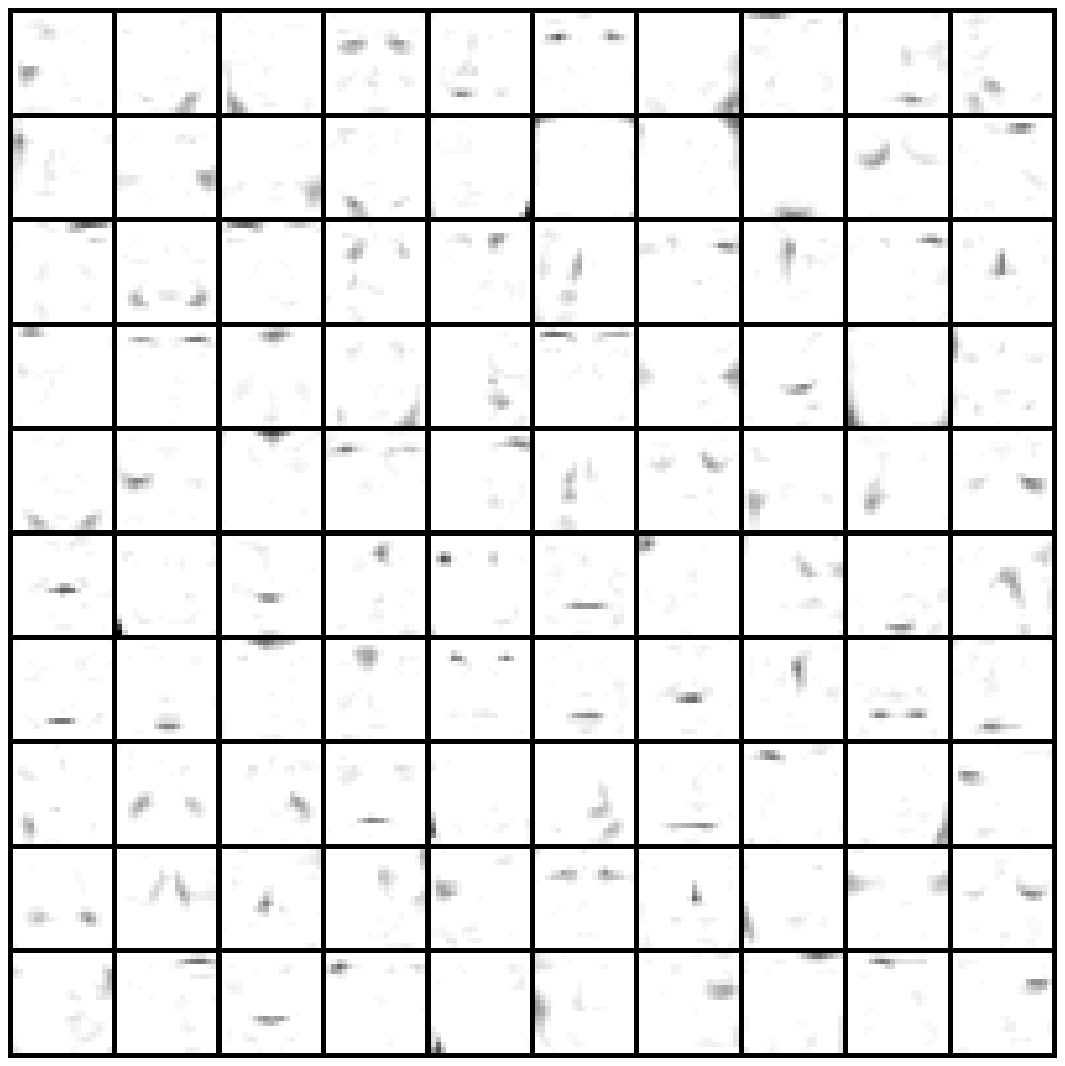}}\caption{Receptive fields learned from the CBCL face image database using RBM,
NMF and $\protect\model$ on Figs.~(\ref{fig:exp_cbcl_filter_rbm},\ref{fig:exp_cbcl_filter_nmf},\ref{fig:exp_cbcl_filter_nrbm}).
Darker pixels show larger weights.\label{fig:experiment_CBCL_filter}}
\end{figure}

For the CBCL dataset, the facial parts (eyes, mouth, nose, eyebrows
etc.) uncovered by $\model$ (Fig.~\ref{fig:exp_cbcl_filter_nrbm})
are visually interpretable along the line with classical NMF \cite{lee_etal_nature99_nmf}
(Fig.~\ref{fig:exp_cbcl_filter_nmf}). The RBM, on the other hand,
produces global facial structures (Fig.~\ref{fig:exp_cbcl_filter_rbm}).
On the more challenging facial set with higher variation such as the
ORL (cf. Sec.~\ref{sec:intro}), NMF fails to produce parts-based
representation (Fig.~\ref{fig:exp_ORL_filter_NMF}), and this is
consistent with previous findings \cite{hoyer_jmlr04_nmf}. In contrast,
the $\model$ is still able to learn facial components (Fig.~\ref{fig:exp_ORL_filter_NRBM}).

\begin{figure}
\noindent \centering{}\subfloat[$\alpha=0.001$]{\noindent \centering{}\includegraphics[width=0.3\textwidth]{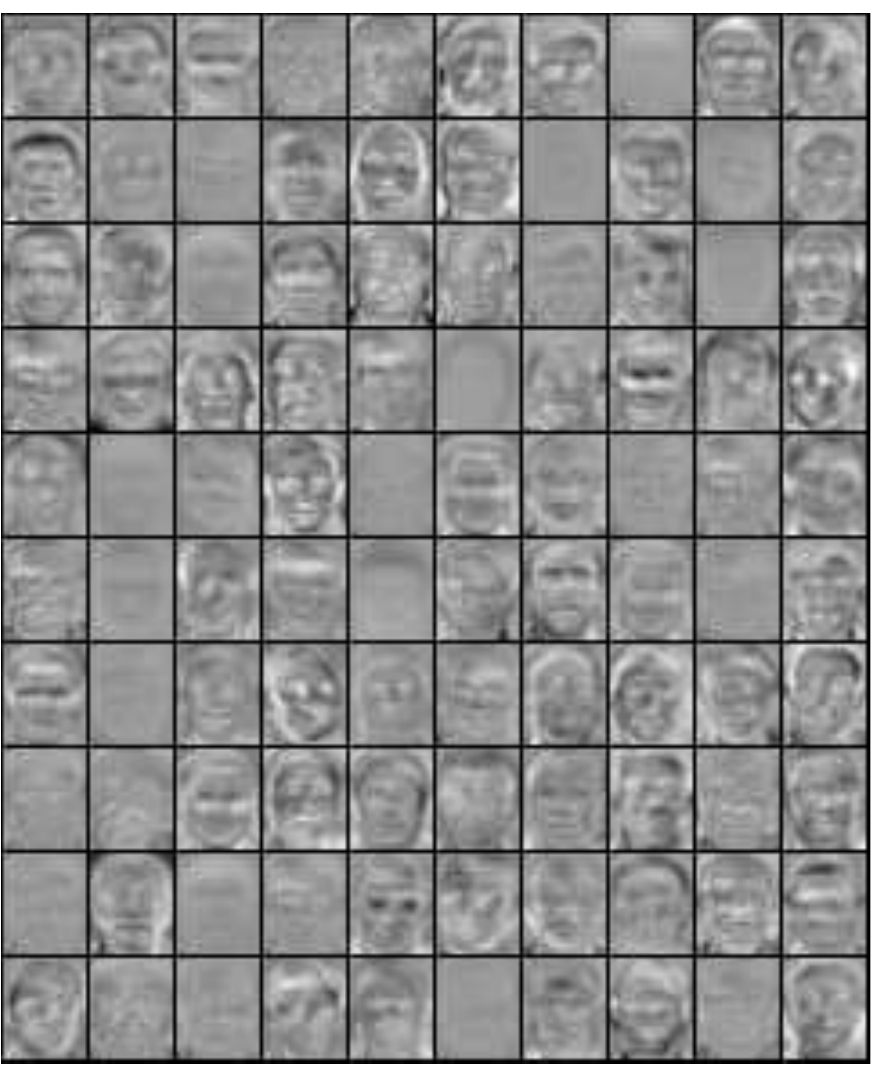}}\hspace{0.02\textwidth}\subfloat[$\alpha=0.01$]{\noindent \centering{}\includegraphics[width=0.3\textwidth]{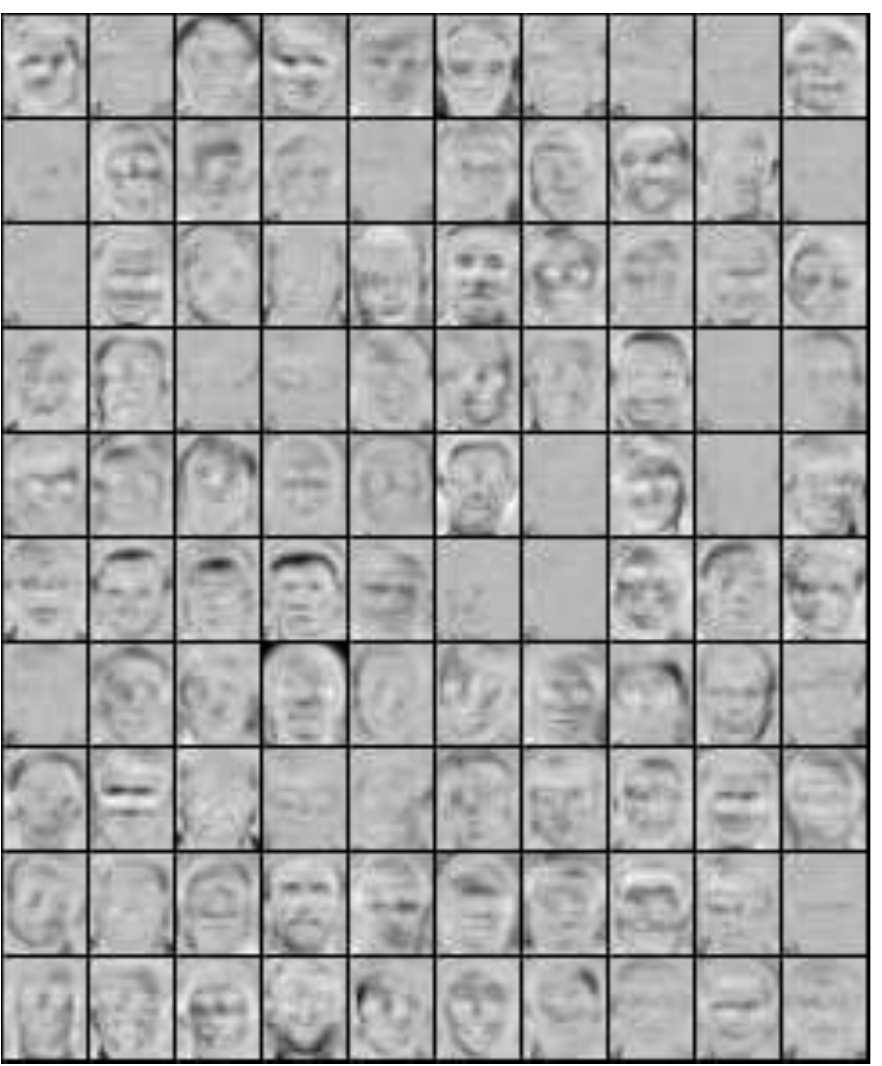}}\hspace{0.02\textwidth}\subfloat[$\alpha=0.1$]{\noindent \centering{}\includegraphics[width=0.3\textwidth]{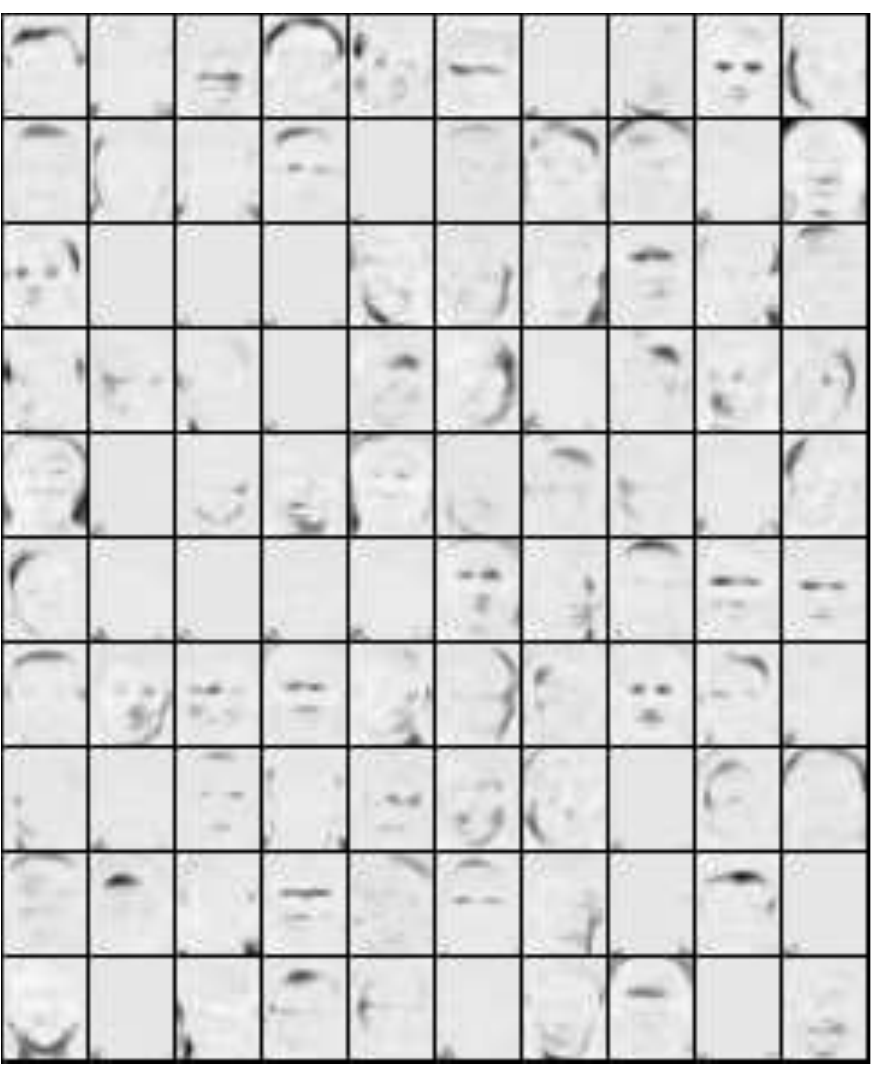}}\caption{Receptive fields learned from the ORL face image database using $\protect\model$
with varied barrier costs. The barrier cost $\alpha$ is tightened
up from left to right. Darker pixels indicate larger weights.\label{fig:exp_orl_filter_alpha}}
\end{figure}

The capacity to decompose data in $\model$ is controlled by a single
hyperparameter $\alpha$. As shown in Fig.~\ref{fig:exp_orl_filter_alpha},
there is a smooth transition from the holistic decomposition as in
standard RBM (when $\alpha$ is near zero) to truly parts-based representations
(when $\alpha$ is larger).

\subsubsection{Dead factors and dimensionality estimation\label{sub:exp_parts_dead}}

We now examine the ability of $\model$ to estimate the intrinsic
dimensionality of the data, as discussed in Section~\ref{sub:nrbm_intrinsic_dim}.
We note that by ``dimensionality'' we roughly mean the degree of
variations, not strictly the dimension of the data manifold. This
is because our latent factors are discrete binary variables, and thus
they may be less flexible than real-valued coefficients.

\begin{figure}
\noindent \centering{}\includegraphics[width=0.6\textwidth]{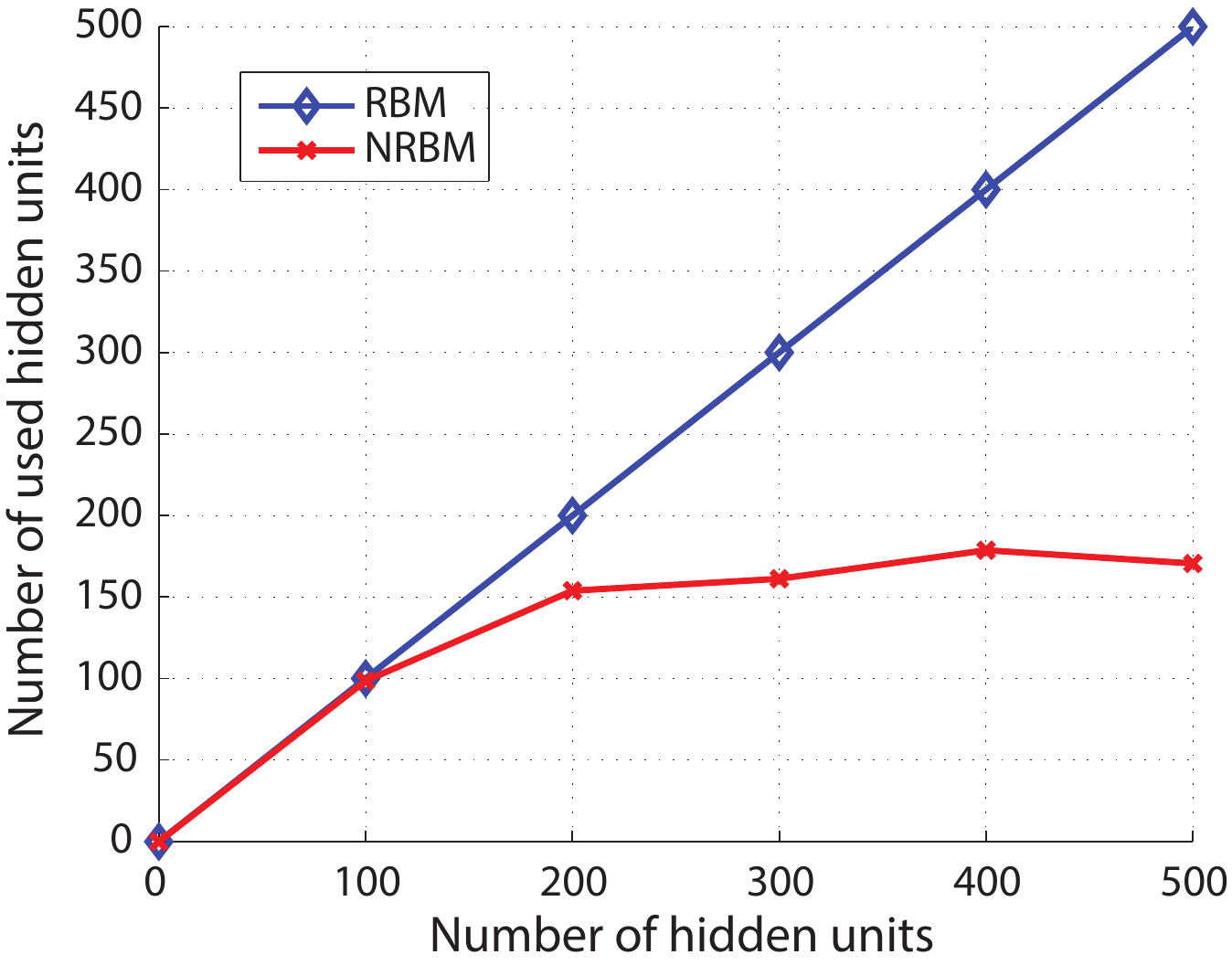}\caption{The numbers of used hidden units with different hidden layer sizes
of RBM vs $\protect\model$.\label{fig:exp_mnist_dead}}
\end{figure}

For that purpose, we compute the number of dead or unused hidden units.
The hidden unit $k$ is declared ``dead'' if the normalized $\ell_{1}$-norm
of its connection weight vector is lower than a threshold $\tau$:
$\left|\bw_{\bigcdot k}\right|_{1}\N^{-1}\leq\tau$, where $\N$ is
the dimension of the original data. We also examine the hidden biases
which, however, do not cause dead units in this case. In Fig.~\ref{fig:exp_mnist_dead},
the number of used hidden units is plotted against the total number
of hidden units $\K$ by taking the average over a set of thresholds
($\tau\in\left\{ 0.01;0.02;...;0.06\right\} $). With the $\model$,
the number of hidden units which explicitly represents the data saturates
at about $150$ whilst all units are used by the RBM.

\subsubsection{Semantic features discovering on text data\label{sub:exp_parts_text}}

The next experiment investigates the applicability of the $\model$
on decomposing text into meaningful ``parts'', although this notion
does not convey the same meaning as those in vision. This is because
the nature of text may be more complex and high-level, and it is hard
to know whether the true nature of word usage is additive. Following
literature in topic modeling (e.g., cf. \cite{blei_etal_jmlr03_lda}),
we start from the assumption that there are latent themes that govern
the choice of words in a particular document. Our goal is to examine
whether we can uncover such themes for each document, and whether
the themes are corresponding to semantically coherent subset of words.

Using the TDT2 corpus, we learn the $\model$ from the data and examine
the mapping weight matrix $\bW$. For each latent factor $k$, the
entry to column $\bw_{\bigcdot k}$ reflects the association strength
of a particular word with the factor, where zero entry means distant
relation. Table~\ref{tab:exp_parts_text_topic} presents four noticeable
semantic features discovered by our model. The top row lists the top
$15$ words per feature, and the bottom row plots the distribution
of association strengths in decreasing order. It appears that the
words under each feature are semantically related in a coherent way.
\begin{table}
\noindent \centering{}\resizebox{0.95\textwidth}{!}{
\begin{tabular}{|c|c|c|c|}
\hline 
\textbf{Asian } & \textbf{Current Conflict} & \textbf{1998 } & \textbf{India - }\tabularnewline
\textbf{Economic Crisis} & \textbf{ with Iraq} & \textbf{Winter Olympics} & \textbf{A Nuclear Power?}\tabularnewline
\hline 
\hline 
FINANCIAL & JURY & BUTLER & COURT\tabularnewline
FUND & GRAND & RICHARD & BAN\tabularnewline
MONETARY & IRAQ & NAGANO & TESTS\tabularnewline
INVESTMENT & SEVEN & INSPECTOR & INDIAS\tabularnewline
FINANCE & IRAQI & CHIEF & TESTING\tabularnewline
WORKERS & GULF & OLYMPICS & INDIA\tabularnewline
INVESTORS & BAGHDAD & RISING & SANCTIONS\tabularnewline
DEBT & SADDAM & GAMES & ARKANSAS\tabularnewline
TREASURY & PERSIAN & COMMITTEE & RULED\tabularnewline
CURRENCY & HUSSEIN & WINTER & INDIAN\tabularnewline
RATES & KUWAIT & OLYMPIC & PAKISTAN\tabularnewline
TOKYO & IRAQS & CHAIRMAN & NUCLEAR\tabularnewline
MARKETS & INSPECTOR & JAPANESE & JUDGE\tabularnewline
IMF & STANDOFF & EXECUTIVE & LAW\tabularnewline
ASIAN & BIOLOGICAL & JAKARTA & ARMS\tabularnewline
\hline 
\vspace{-0.001\textheight}
 & \vspace{-0.001\textheight}
 & \vspace{-0.001\textheight}
 & \vspace{-0.001\textheight}
\tabularnewline
\includegraphics[scale=0.2]{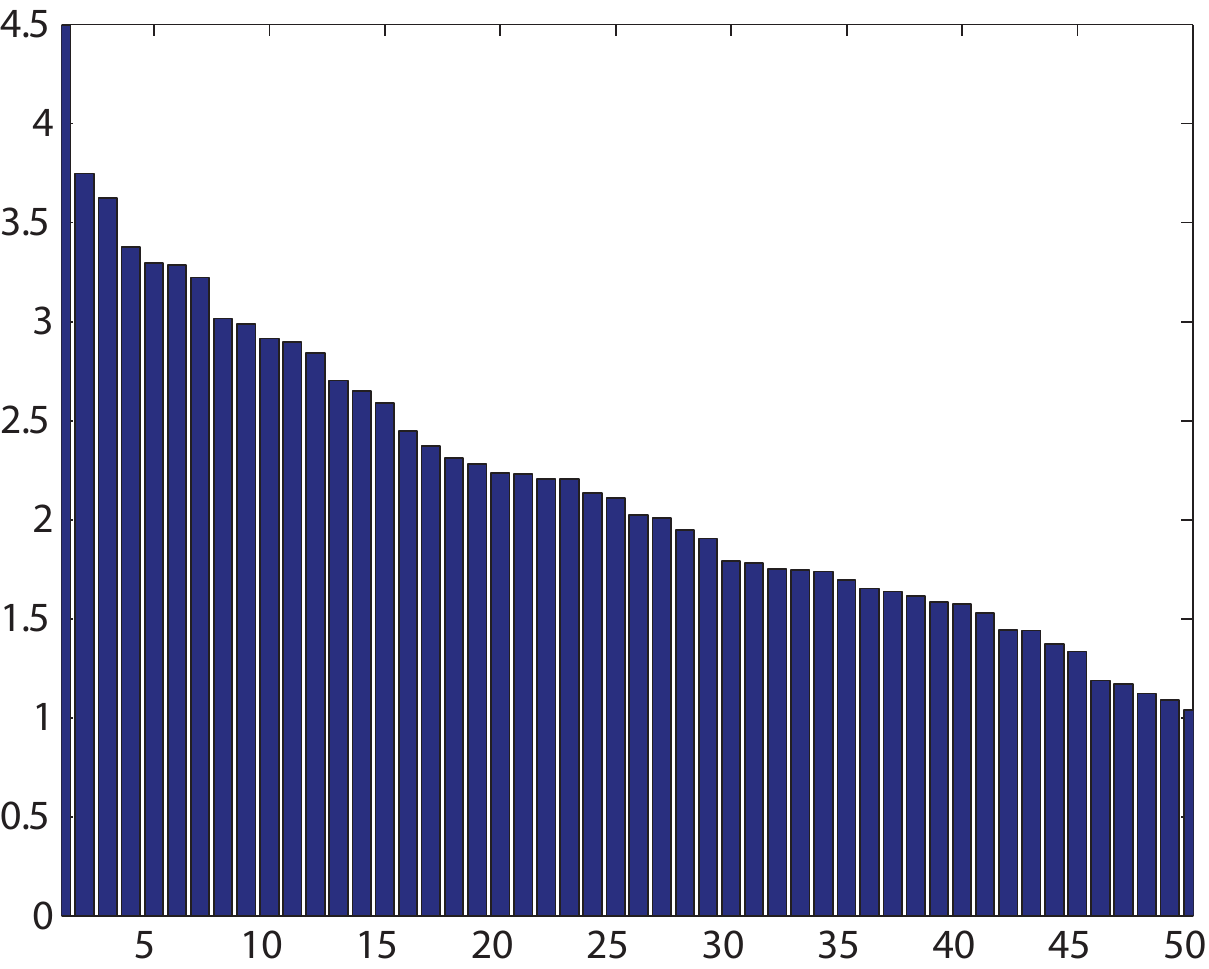} & \includegraphics[scale=0.2]{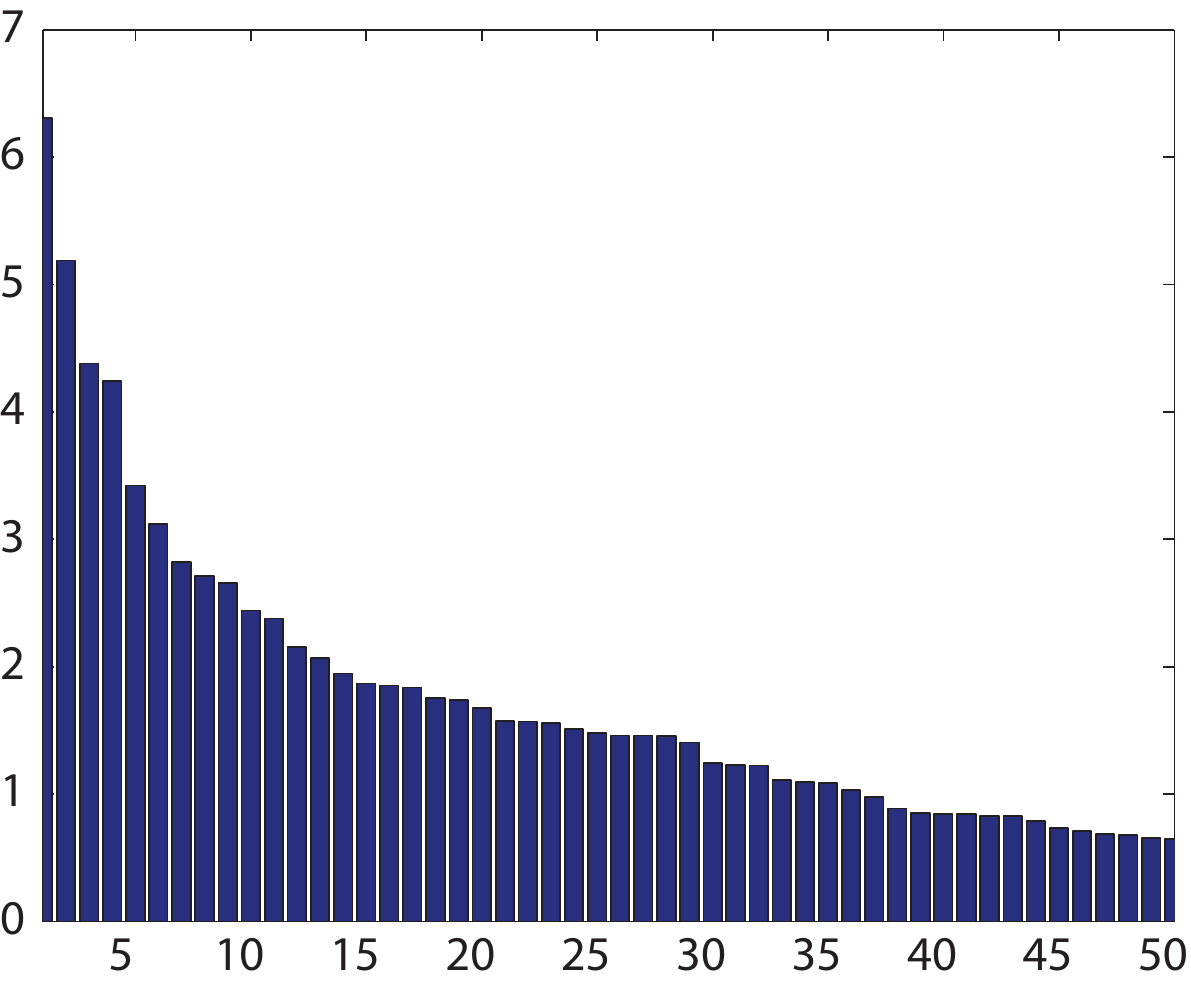} & \includegraphics[scale=0.2]{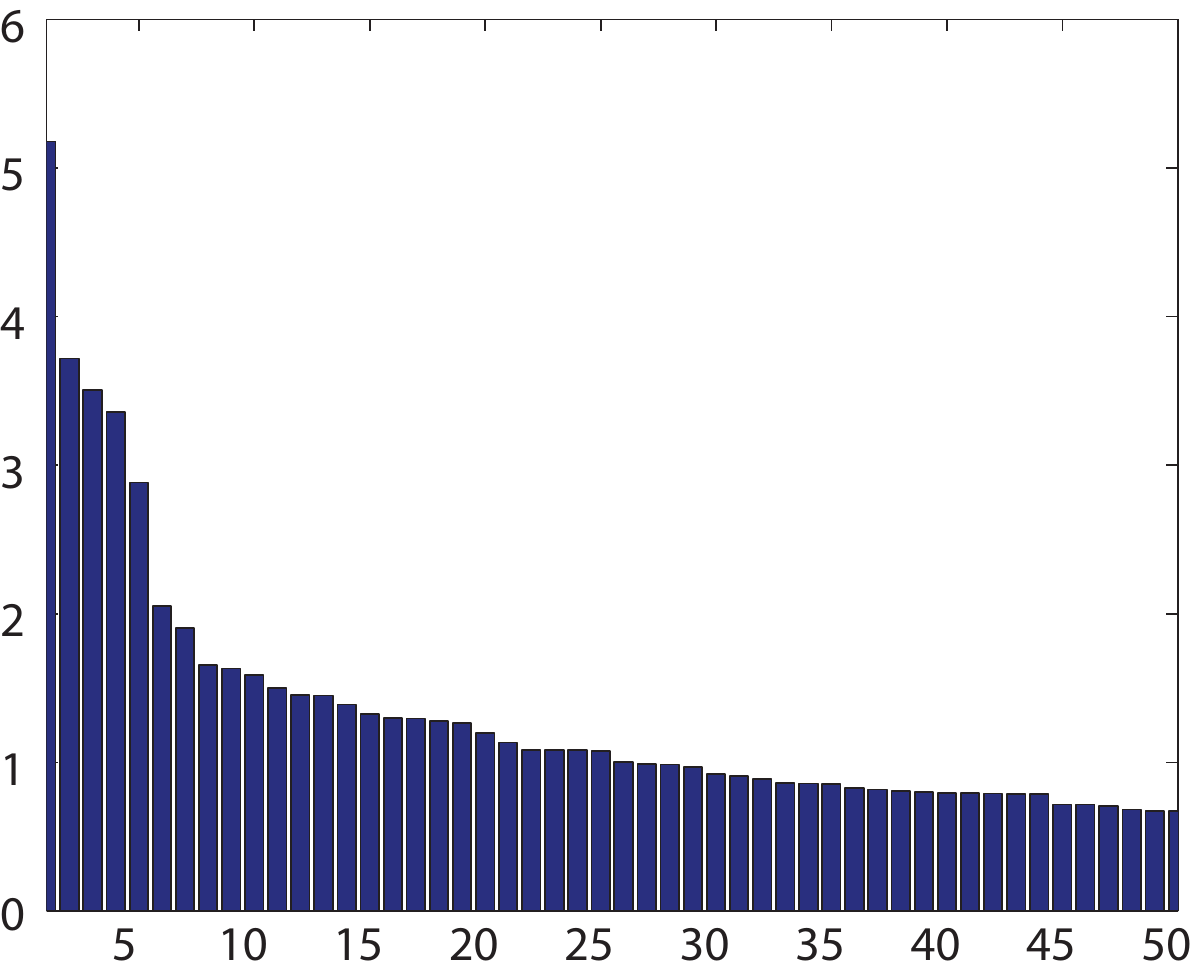} & \includegraphics[scale=0.2]{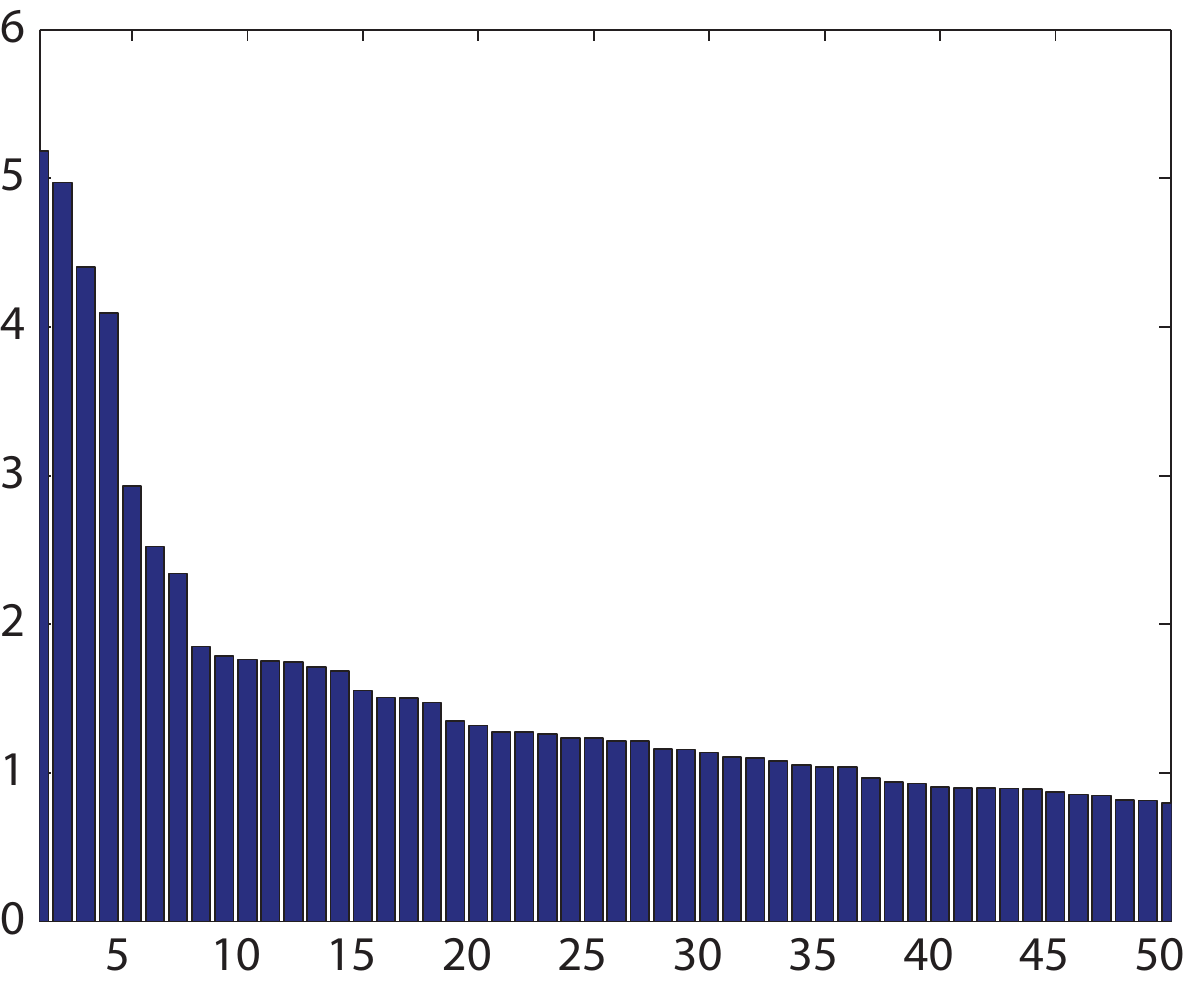}\tabularnewline
\hline 
\end{tabular}}\caption{An example of $4$ most distinguished categories, i.e. economics,
politics, sport and armed conflict associated with top $15$ words
(ranked by their weights) discovered from the TDT2 subset. The charts
at the bottom row illustrate the weight impact of words on the category.
These weights are sorted in descending order.\label{tab:exp_parts_text_topic}}
\end{table}

\subsection{Feature extraction for classification}

Our next target is to evaluate whether the ability to decompose data
into parts and to disentangle factors of variation could straightforwardly
translate into better predictive performance. Although the $\model$
can be easily turned into a nonnegative neural network and the weights
are tuned further to best fit the supervised setting (e.g., cf. \cite{hinton_salakhutdinov_sci06_reducing}),
we choose not to do so because our goal is to see if the decomposition
separates data well enough. Instead we apply standard classifiers
on the learned features, or more precisely the hidden posteriors.

The first experiment is with the MNIST, the $500$ factors have been
previously learned in Section~\ref{sub:exp_parts_decom} and Fig.~\ref{fig:exp_mnist_filter}.
Support vector machines (SVM, with Gaussian kernels, using the LIBSVM
package \cite{chang_etal_tist11_libsvm}) and $k$-nearest neighbors
($k$NN, where $k=4$, with cosine similarity measures) are used as
classifiers. For comparison, we also apply the same setting to the
features discovered by the NMF. The error rate on test data is reported
in Table~\ref{tab:exp_mnist_clf}. It can be seen that (i) compared
to standard RBM, the nonnegativity constraint used in $\model$ does
not lead to a degradation of predictive performance, suggesting that
the parts are also indicative of classes; and (ii) nonlinear decomposition
in $\model$ can lead to better data separation than the linear counterpart
in NMF.
\begin{table}
\noindent \centering{}\resizebox{0.4\textwidth}{!}{
\begin{tabular}{|c|c|c|}
\cline{2-3} 
\multicolumn{1}{c|}{} & SVM & $4$-NN\tabularnewline
\hline 
RBM & \textbf{1.38} & 2.74\tabularnewline
\hline 
NMF & 3.25 & 2.64\tabularnewline
\hline 
NRBM & 1.4 & \textbf{2.34}\tabularnewline
\hline 
\end{tabular}}\caption{The classification errors (\%) on testing data of MNIST dataset.\label{tab:exp_mnist_clf}}
\end{table}

\begin{figure}
\noindent \centering{}\subfloat[2D projection of hidden posteriors.\label{fig:exp_tdt2_tsne}]{\noindent \centering{}\includegraphics[width=0.49\textwidth]{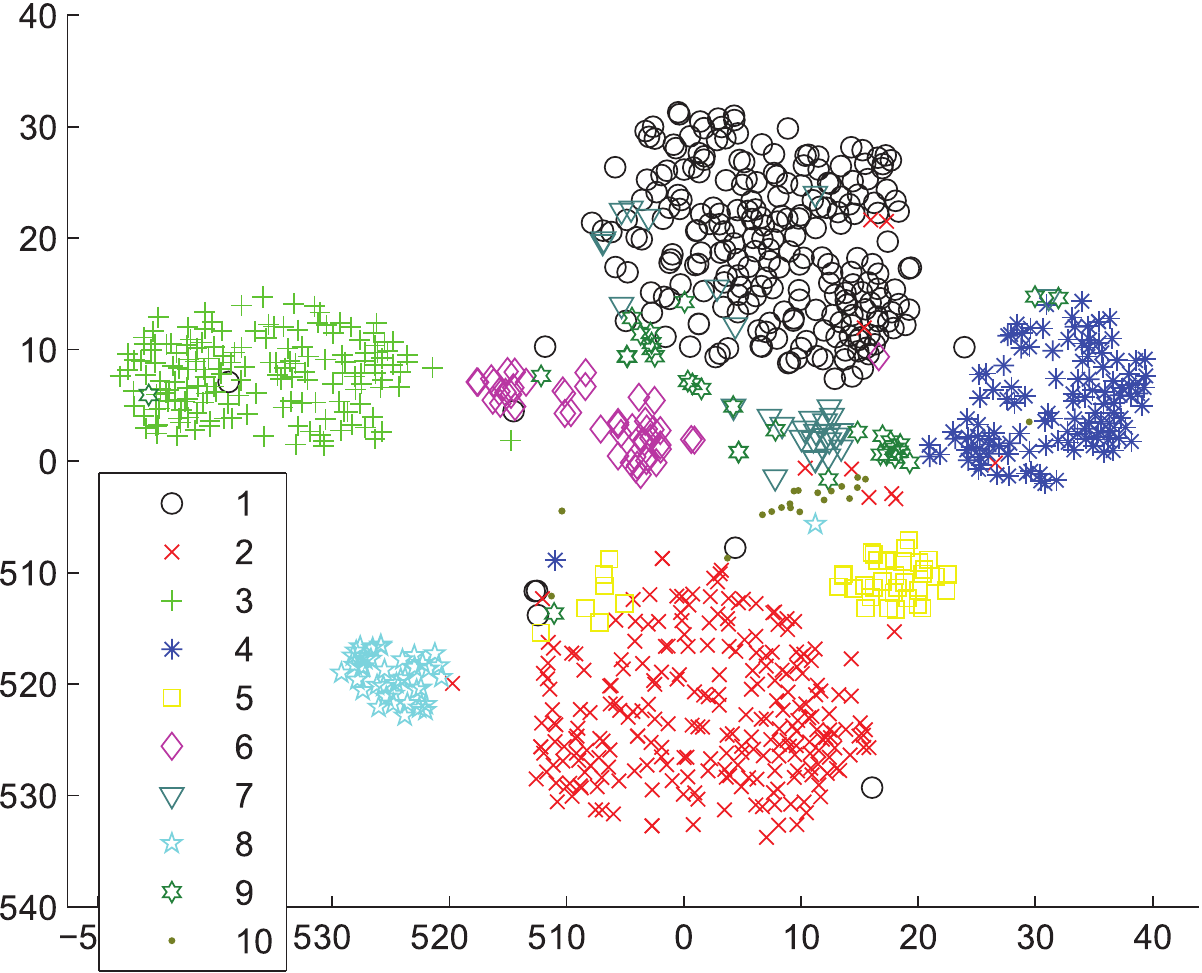}}\hfill{}\subfloat[Classification errors $\left(\%\right)$ on TDT2 corpus.\label{fig:exp_tdt2_clf}]{\noindent \centering{}\includegraphics[width=0.49\textwidth]{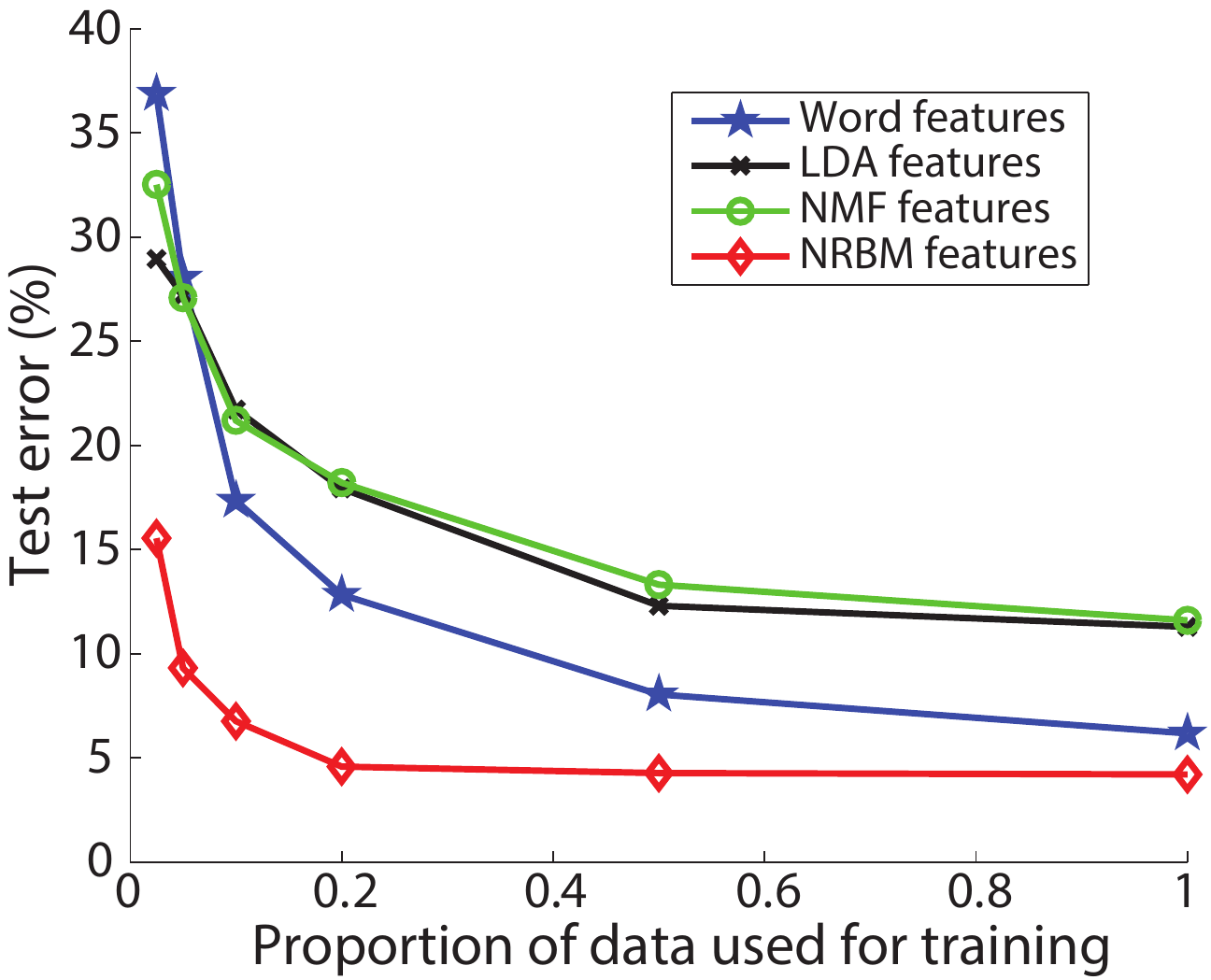}}\caption{An example visualization of 10 categories and classification performance
on TDT2 text corpus. On Fig.~\ref{fig:exp_tdt2_tsne}, t-SNE projection
\cite{van_hinton_jmlr08_tsne} performs on $100$ higher representations
of documents mapped using the $\protect\model$. Categories are labeled
using the ground truth. (Best viewed in colors). Fig.~\ref{fig:exp_tdt2_clf}
represents the classification results for different proportions of
training data.\label{fig:NRBM_exp_tdt2}}
\end{figure}

The second experiment is on the text data TDT2. Unlike images, words
are already conceptual and thus using standard bag-of-words representation
is often sufficient for many classification tasks. The question is
therefore whether the thematic features discovered by the $\model$
could further improve the performance, since it has been a difficult
task for topic models such as LDA (e.g., cf. experimental results
reported in \cite{blei_etal_jmlr03_lda}). To get a sense of the capability
to separate data into classes without the class labels, we project
the $100$ hidden posteriors onto 2D using t-SNE\footnote{Note that the t-SNE does not do clustering, it only reduces the dimensionality
into 2D for visualization while still try to preserve the local properties
of the data.} \cite{van_hinton_jmlr08_tsne}. Fig.~\ref{fig:exp_tdt2_tsne} depicts
the distribution of documents, where class information is only used
for visual labeling. The separation is overall satisfactory.

For the quantitative evaluation, the next step is to run classifiers
on learned features. For comparison, we also use those discovered
by NMF and LDA. For all models, $100$ latent factors are used, and
thus the dimensions are reduced $10$-fold. We split TDT2 text corpus
into $80\%$ for training and $20\%$ for testing. We train linear
SVMs\footnote{SVM with Gaussian kernels did not perform well.} on
all word features and low-dimensional representations provided by
LDA, NMF and $\model$ with various proportions of training data.
Fig.~\ref{fig:exp_tdt2_clf} shows the classification errors on testing
data for all methods. The learned features of LDA and NMF improve
classification performance when training label is limited. This is
expected because the learned representations are more compact and
thus less prone to overfitting. However, as more labels are available,
the word features catch up and eventually outperform those by LDA/NMF.
Interestingly, this difficulty does not occur for the features learned
by the $\model$, although it does appear that the performance saturates
after seeing 20\% of training labels. Note that this is significant
given the fact that computing learned representations in $\model$
is very fast, requiring only a single matrix-vector multiplication
per document.

\subsection{Stabilizing linear predictive models\label{sub:exp_featstab}}

In this experiment, we evaluate the stability of feature selection
of our framework introduced in Section~\ref{sec:nrbm_stab}. More
specifically, we assess the discovered risk factors of heart failure
patients by predicting the future presence of their readmission (i.e.,
$\y_{m}$ in Eq.~(\ref{eq:NRBM_modelstab_pdf})) at a certain assessment
point given their history. It is noteworthy that this is more challenging
as the EMR data contain rich information, are largely temporal, often
noisy, irregular, mixed-type, mixed-modality and high-dimensional
\cite{luo_etal_sdm2012_sor}. In what follows, we present the experiment
setting, evaluation protocol and results.

\subsubsection{Temporal validation}

\begin{figure}
\noindent \centering{}\includegraphics[width=0.8\textwidth]{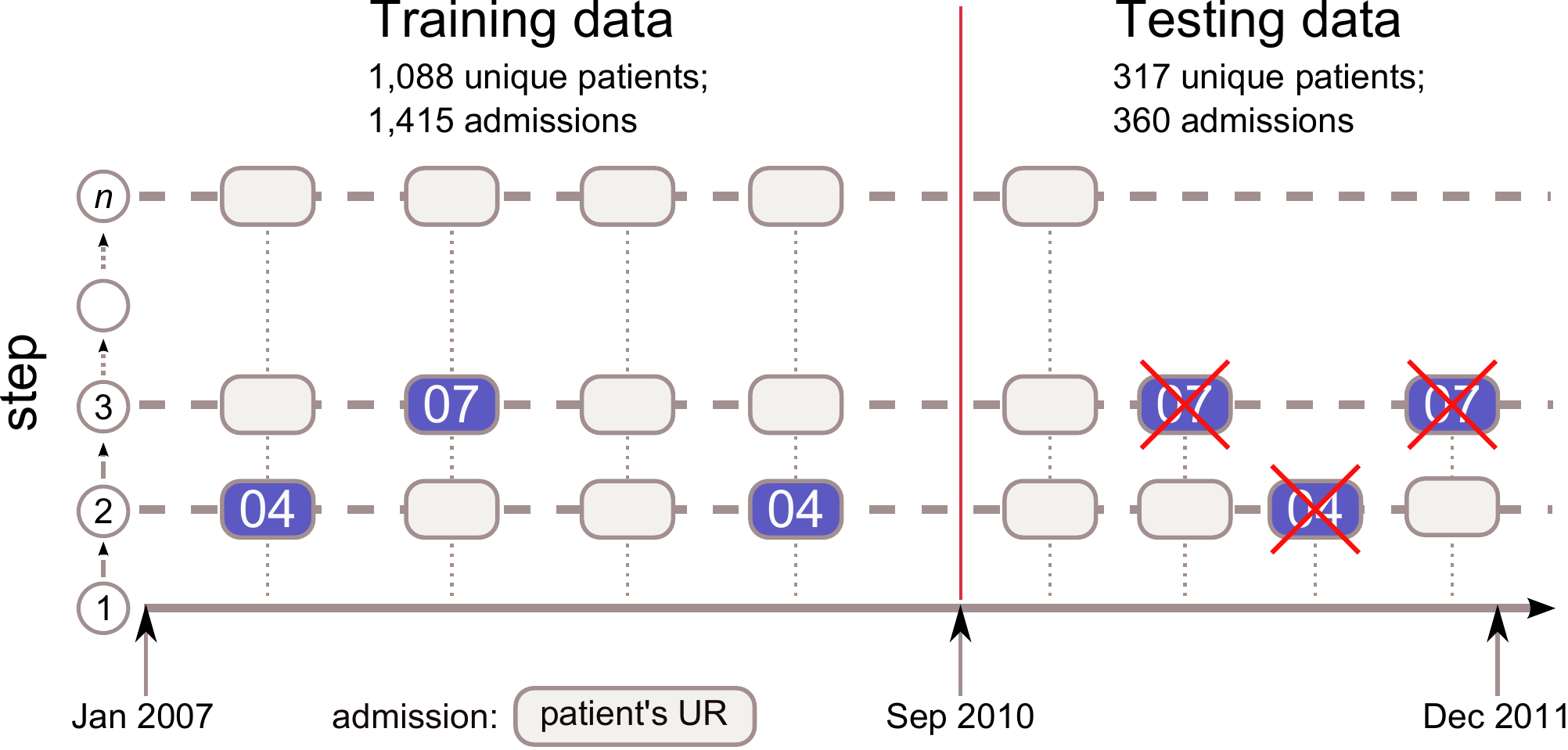}\caption{Data splitting process. The cohort was divided into two parts with
before September 2010 for training and after that for testing. All
admissions of patients in the testing set who are present in training
set were then removed to achieve non-overlapping property.\label{fig:NRBM_exp_data_splitting}}
\end{figure}

We derive the cohort into training and testing data to validate the
predictive performance of our proposed model. Two issues that must
be addressed during the splitting process are: learning the past and
predicting the future; ensuring training and testing sets completely
separated. Here we use a temporal checkpoint to divide the data into
two parts. More specifically, we gather admissions which have discharge
dates before September 2010 to form the training set and after that
for testing. Next we specify the set of unique patients in the training
set. We then remove all admissions of such patients in the testing
set to guarantee no overlap between two sets. Finally, we obtain $1,088$
unique patients with $1,415$ admissions in the training data and
$317$ patients with $360$ admissions in the testing data. The removing
steps and resulting datasets are illustrated in Fig.~\ref{fig:NRBM_exp_data_splitting}.
Our model is then learned using training data and evaluated on testing
data.

\subsubsection{Evaluation protocol}

We use Jaccard index \cite{real_vargas_1996_probabilistic} and consistency
index \cite{kuncheva_aia07_stability} to measure the stability of
feature selection process. The Jaccard index, also known as the Jaccard
similarity coefficient, naturally considers both similarity and diversity
to measure how two feature sets are related. The consistency index
supports feature selection in obtaining several desirable properties,
i.e. monotonicity, limits and correction for chance.

We trained our proposed model by running $\M=10$ bootstraps and obtained
a list of feature sets $\mathcal{S}=\left\{ \S_{1},\S_{2},...,\S_{\M}\right\} $
where $\S_{i}$ is a subset of original feature set $\bv$. Note that
the cardinalities are: $\left|\S_{i}\right|=\T$ and $\left|\v\right|=\K$
with the condition: $\T\leq\K$. Considering a pair of subsets $\S_{i}$
and $\S_{j}$, the pairwise consistency index $C\left(\S_{i},\S_{j}\right)$
is defined as:
\begin{align*}
C\left(\S_{i},\S_{j}\right) & =\frac{\R\K-\T^{2}}{\T(\K-\T)}
\end{align*}
in which $\left|\S_{i}\cap\S_{j}\right|=\R$. Taking the average of
all pairs, the overall consistency index is:
\begin{align}
C & =\frac{2}{\M(\M-1)}\overset{_{\M-1}}{\underset{_{i=1}}{\sum}}\overset{_{\M}}{\underset{_{j=i+1}}{\sum}}C(\S_{i},\S_{j})\label{eq:NRBM_exp_ci}
\end{align}
Whilst most similarity indices prefer large subsets, $C$ provides
consistency around zero for any number of features $\T$ \cite{kuncheva_aia07_stability}.
The consistency index is bounded in $\left[-1,+1\right]$.

Jaccard index measures similarity as a fraction between cardinalities
of intersection and union feature subsets. Given two feature sets
$\S_{i}$ and $\S_{j}$, the pairwise Jaccard index $J\left(\S_{i},\S_{j}\right)$
reads:
\begin{align*}
J(\S_{i},\S_{j}) & =\frac{\left|\S_{i}\cap\S_{j}\right|}{\left|\S_{i}\cup\S_{j}\right|}
\end{align*}
The Jaccard index evaluating all $\M$ subsets was computed as follows:
\begin{align}
J & =\frac{2}{\M(\M-1)}\overset{_{\M-1}}{\underset{_{i=1}}{\sum}}\overset{_{\M}}{\underset{_{j=i+1}}{\sum}}J(\S_{i},\S_{j})\label{eq:NRBM_exp_ji}
\end{align}
Jaccard index is bounded in $\left[0,1\right]$.

For prediction, we average the weights of $10$ models learned after
bootstrapping to obtain the final model. Then the final model performs
prediction on testing data. The threshold $0.5$ is used to decide
the predicted outcomes from predicted probabilities. Finally the performances
are evaluated using measures of sensitivity (recall), specificity,
precision, F-measure and area under the ROC curve (AUC) score with
confidence intervals based on Mann-Whitney statistic \cite{birnbaum_etal_1956_use}.

\subsubsection{Results\label{sub:exp_modelstab_res}}

Section~\ref{sec:related_work} shows that the basis matrix of NMF
plays the same role as NRBM's connection weights. Using the same derivation
of $\model$ in Section~\ref{sec:nrbm_stab}, it is straightforward
to obtain the weight vector $\bbarw$ in Eq.~(\ref{eq:NRBM_modelstab_conjugated_w})
for the NMF. Thus we can compare the stability results of our proposed
model against those of NMF. In total, we recruit three baselines:
lasso, RBM and NMF. The numbers of hidden units, latent factors of
RBM, $\model$ and NMF are $200$.

\begin{figure}
\noindent \begin{centering}
\hspace{0.05\textwidth}\includegraphics[width=0.75\textwidth]{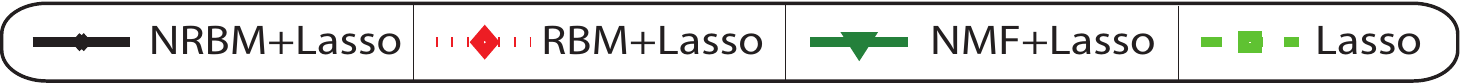}
\par\end{centering}

\noindent \centering{}\subfloat[Consistency index.\label{fig:NRBM_exp_featstab_ci}]{\noindent \centering{}\includegraphics[width=0.49\textwidth]{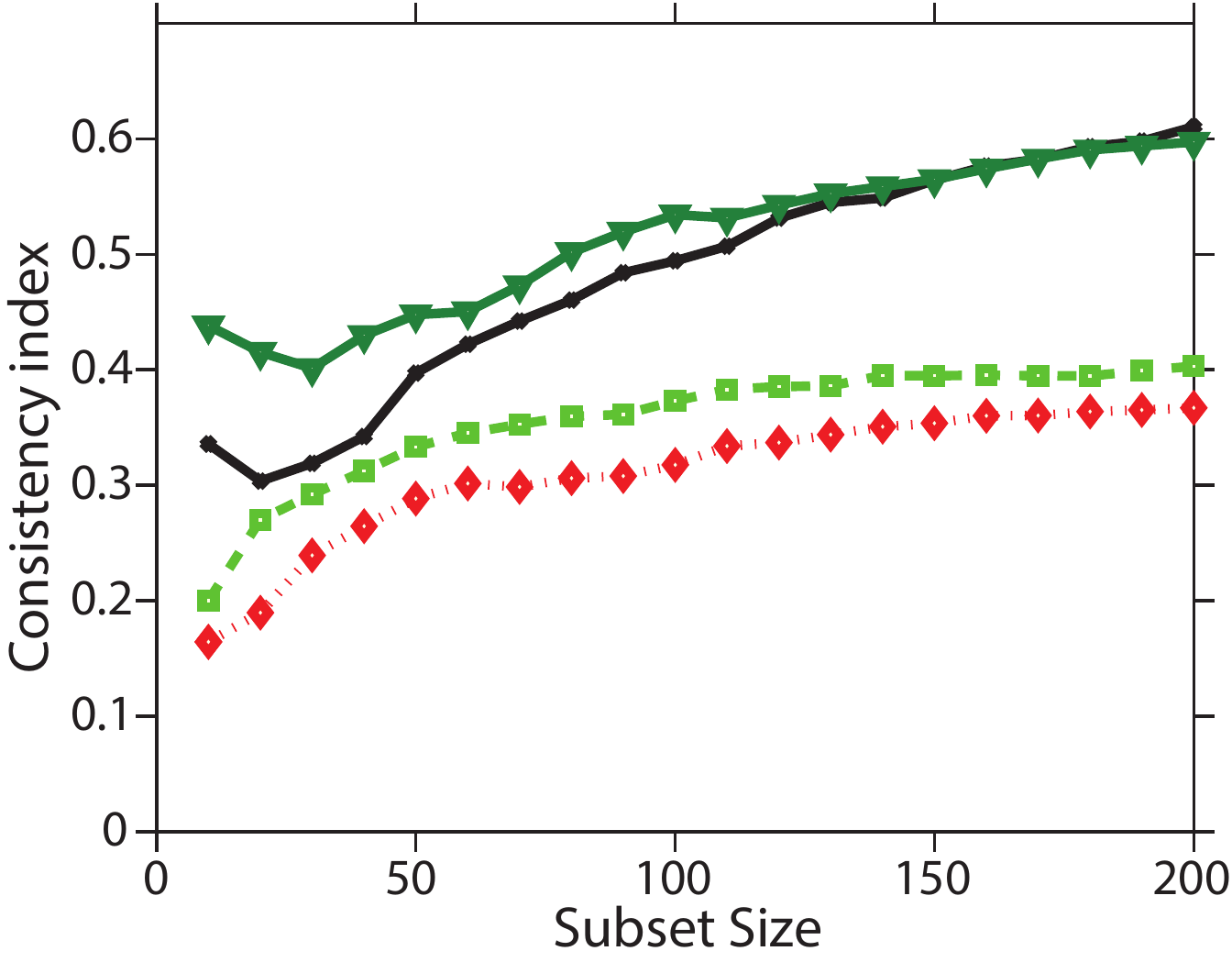}}\hfill{}\subfloat[Jaccard index.\label{fig:NRBM_exp_featstab_ji}]{\noindent \centering{}\includegraphics[width=0.49\textwidth]{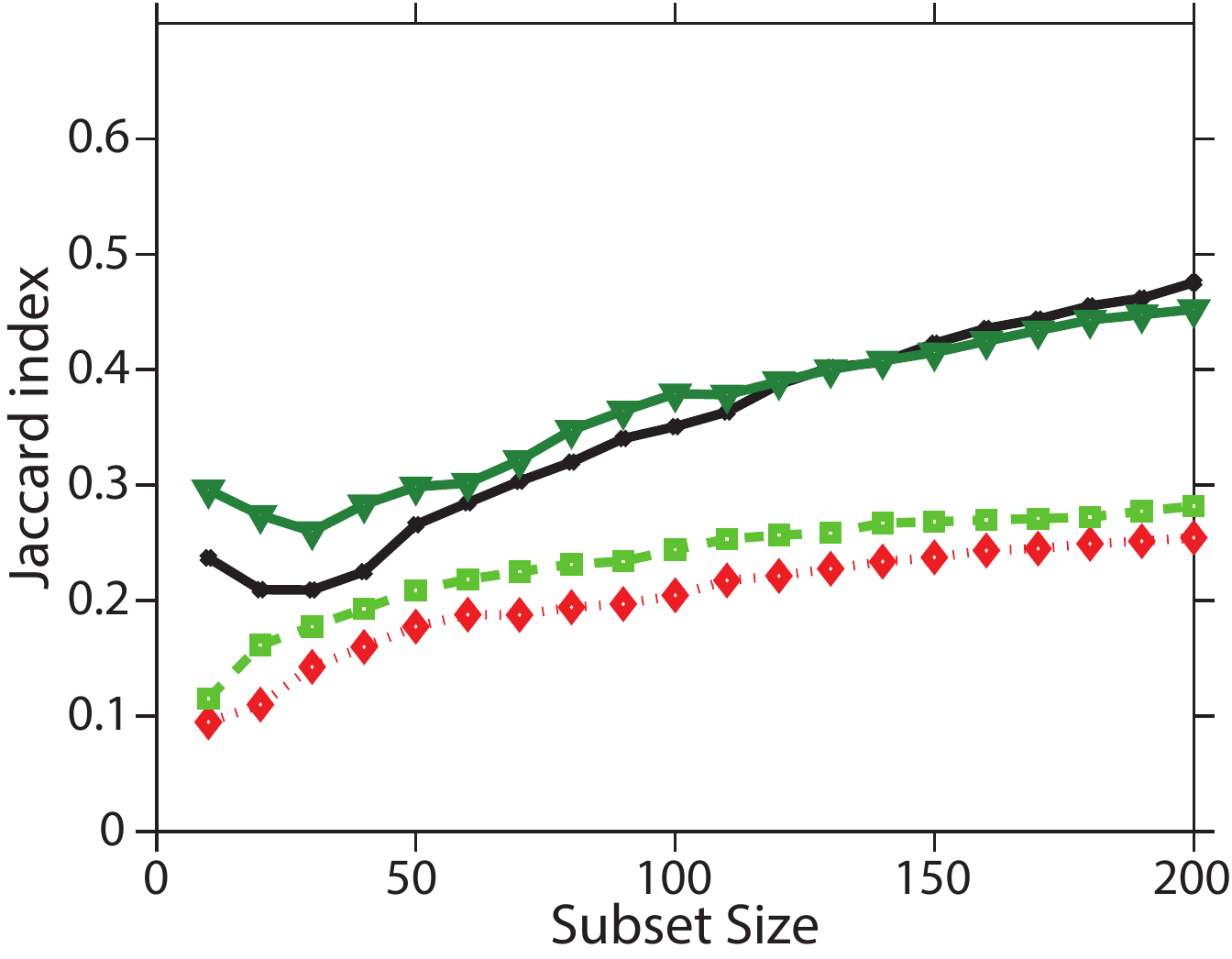}}\caption{Feature selection stability measured by the consistency index (Fig.~\ref{fig:NRBM_exp_featstab_ci})
and Jaccard index (Fig.~\ref{fig:NRBM_exp_featstab_ji}) for 6-month
heart failure prediction.\emph{\label{fig:exp_featstab}}}
\end{figure}

Table.~\ref{tab:exp_featstab_predres} reports the prediction performances
and Fig.~\ref{fig:exp_featstab} illustrates the stability results
for different subset sizes. Overall, the $\model$ followed by lasso
achieve better AUC score than the RBM and NMF followed by lasso and
worse yet acceptable ($1.8\%$ lower) than the sole lasso. However,
the $\model$ followed by the lasso outperform the sole lasso and
RBM$+$Lasso with large margins of consistency and Jaccard indices.
The worst stabilities of RBM$+$Lasso are expected because the standard
RBM is not designed with properties that promote a steady group of
features. Comparing with the NMF followed by lasso, the $\model$
performs worse at first but then catches up when the subset size reaches
about $150$ and slightly better after that.
\begin{table}
\noindent \centering{}\resizebox{0.95\textwidth}{!}{
\begin{tabular}{|c|c|c|c|c|c|}
\hline 
Method & Sens./Rec. & Spec. & Prec. & F-mea. & AUC {[}CI{]}\tabularnewline
\hline 
\hline 
Lasso & $0.6137$ & $0.5903$ & $0.4833$ & $0.5404$ & $0.6213\left[0.5644,0.6781\right]$\tabularnewline
\hline 
RBM$+$Lasso & $0.5726$ & $0.5551$ & $0.3944$ & $0.4671$ & $0.6001\left[0.5425,0.6577\right]$\tabularnewline
\hline 
NMF$+$Lasso & $0.5224$ & $0.5536$ & $0.5833$ & $0.5512$ & $0.5690\left[0.5107,0.6274\right]$\tabularnewline
\hline 
NRBM$+$Lasso & $0.5816$ & $0.5587$ & $0.4011$ & $0.4748$ & $0.6083\left[0.5508,0.6659\right]$\tabularnewline
\hline 
\end{tabular}}\caption{The prediction performance of the $\protect\model$ and baselines.\label{tab:exp_featstab_predres}}
\end{table}

\section{Discussion\label{sec:related_work}}

\subsection{Representation learning}

Representation learning has recently become a distinct area in machine
learning after a long time being underneath the subdomain of ``feature
learning''. This is evidenced through the recent appearance of the
international conference on learning representation (ICLR\footnote{\href{http://www.iclr.cc/doku.php}{http://www.iclr.cc/doku.php}}),
following a series of workshops in NIPS/ICML \cite{bengio_etal_pami13_representation}.
Existing approaches of learning representation can be broadly classified
into three categories: learning kernels, designing hand-crafted features,
and learning high-level representations.

In kernel learning approach, the algorithms may use fixed generic
kernels such as Gaussian kernel, design kernels based on domain knowledge,
or learn combined kernels from multiple base kernels \cite{scholkopf_etal_mit99_advances}.
These methods often use ``\emph{kernel tricks}'' to substitute the
kernel functions for the mapping functions that transform data into
their new representations on the feature space. Thus the kernel machines
do not define \emph{explicit} mapping functions, and as a result do
not directly produce representations for data.

The second approach is to directly use either raw data or features
extracted from them. These features are extracted from feature selection
frameworks with or without domain knowledge. The original data and
extracted features can be preprocessed by scaling or normalizing,
but not projecting onto other spaces \cite{lowe_ijcv04_sift,dalal_triggs_cvpr05_hog,oliva_torralba_2006_gist,bay_etal_eccv06_surf}.
Normally, the feature selection frameworks are designed by hand. Hand-crafted
feature extraction relies on the design of preprocessing pipelines
and data transformations. This approach, however, suffers from two
significant drawbacks. First, it is labor intensive and normally requires
exhaustive prior knowledge. Thus only domain experts can design good
features. Second, the feature engineering process may create more
data types, especially for complex data, leading to more challenges
for fundamental machine learning methods.

The last approach is to automatically learn parametric maps that transform
input data into high-level representations. The models that learn
such representations typically fall into two classes: one is non-probabilistic
models, the other is probabilistic latent variable models. The common
aim of the methods in the first class is to learn direct encoding,
i.e., parametric maps from input data to their new representations.
The classic linear techniques are principal component analysis (PCA)
\cite{jolliffe_springer86_pca}, latent semantic indexing (LSI) \cite{deerwester_etal_jasis90_lsi},
and independent component analysis (ICA) \cite{hyvarinen_oja_nn00_independent}.
These methods support capturing data regularity, discovering latent
structures and latent semantics of data. They, however, face difficulty
in modeling data which follow a set of probabilistic distributions.
This drawback can be overcome by the second class of probabilistic
models.

Probabilistic latent variable models often consist of observed variables
$\bx$ representing data and latent variables $\bh$ which coherently
reflect the distributions, regularities, structures and correlations
of data. The common idea is to define a joint probability $p\left(\bx,\bh\right)$
over the joint space of observed and latent variables. The learning
is to maximize the likelihood of training data. Once the model has
been learned, the latent representation of the data is obtained by
inferring the posterior probability $p\left(\bh\gv\bx\right)$. Most
methods in this part are probabilistic versions of the ones in the
first class including probabilistic PCA \cite{tipping_etal_jrss99_pPCA},
probabilistic LSI \cite{hofmann_uai99_pLSI}, and Bayesian nonparametric
factor analysis \cite{paisley_etal_icml09_bnfa}. As a fully generative
two-layer model, our proposed model can also be categorized into this
class.

\subsection{Nonnegative data modeling}

Our work was partly motivated by the capacity of nonnegative matrix
factorization (NMF) \cite{lee_etal_nature99_nmf} to uncover parts-based
representations. Given a nonnegative data matrix $\bV\in\mathbb{R}^{\N\times\M}$,
the NMF attempts to factorize into two low-rank real-valued nonnegative
matrices, the basis $\bW\in\mathbb{R}^{\N\times\K}$ and the coefficient
$\bH\in\mathbb{R}^{\K\times\M}$, i.e., $\bV\approx\bW\bH$. Thus
$\bW$ plays the same role as NRBM's connection weights, and each
column of $\bH$ assumes the ``latent factors''. However, it has
been pointed out that unless there are appropriate sparseness constraints
or certain conditions, the NMF is not guaranteed to produce parts
\cite{hoyer_jmlr04_nmf}. Our experiment shows, on the contrary, the
$\model$ can still produce parts-based representation when the NMF
fails (Fig.~\ref{fig:exp_ORL_filter}, also reported in \cite{hoyer_jmlr04_nmf}).

On the theoretical side, the main difference is that, in our cases,
the latent factors are stochastic binary that are inferred from the
model, but not learned as in the case of NMF. In fact this seemingly
subtle difference is linked to a fundamental drawback of the NMF:
The learned latent factors are limited to seen data only, and must
be relearned for every new data point. The $\model$, on the other
hand, is a fully generative model in that it can generate new samples
from its distribution, and at the same time, the representations can
be efficiently computed for unseen data (cf. Eq.~(\ref{eq:RBM_hidprob}))
using one matrix operation.

Recently, there has been work closely related to the NMF that does
not require re-estimation on unseen data \cite{lemme_etal_nn12_online}.
In particular, the coefficient matrix $\bH$ is replaced by the mapping
from the data itself, that is $\bH=\boldsymbol{\sigma}\left(\bW^{\top}\bV\right)$,
resulting in the so-called autoencoder structure (AE), that is $\bV\approx\bW\boldsymbol{\sigma}\left(\bW^{\top}\bV\right)$,
where $\boldsymbol{\sigma}(\boldsymbol{x})$ is a vector of element-wise
nonlinear transforms and $\bW$ is nonnegative. A new representation
estimated using $\boldsymbol{\sigma}\left(\bW^{\top}\bV\right)$ now
plays the role of the posteriors in $\model$, although it is non-probabilistic.
The main difference from the $\model$ is that the nonnegative AE
does not model data distribution, and thus cannot generate new samples.
Also, it is still unclear how the new representation could be useful
for classification in general and on non-vision data in particular.

\subsection{Feature discovery}

For the semantic analysis of text, our proposed model is able to discover
plausible thematic features. Compared against those discovered by
topic models such as latent Dirichlet allocation (LDA) \cite{blei_etal_jmlr03_lda},
we found that they are qualitatively similar. We note that the two
approaches are not directly comparable because the notion of association
strength between a latent factor and a word, as captured in the nonnegative
weight $\w_{nk}$, cannot be translated into the properly normalized
probability $P(\v_{n}=1\mid\z_{n}=k)$ as in LDA, where $\z_{n}$
is the topic that generates the word $\v_{n}$. Nevertheless, the
$\model$ offers many advantages over the LDA: (i) the notion that
each document is generated from a subset of themes (or semantic features)
in the $\model$ is an attractive alternative to the setting of topic
distribution as assumed in the LDA (cf. also \cite{griffiths_ghahramani_2005_infinite});
(ii) inference to compute the latent representation given an input
is much faster in the $\model$ with only one matrix multiplication
step, which typically requires an expensive sampling run in the LDA;
(iii) learning in the $\model$ can be made naturally incremental,
whilst estimating parameter posteriors in the LDA generally requires
the whole training data; and (iv) importantly, as shown in our experiments,
classification using the learned representations can be more accurate
with the $\model$.

This work can be considered along the line of imposing structures
on standard RBM so that certain regularities are explicitly modeled.
Our work has focused on nonnegativity as a way to ensure sparseness
on the weight matrix, and consequently the latent factors. An alternative
would be enforcing sparseness on the latent posteriors, e.g., \cite{hinton_springer12_practical}.
Another important aspect is that the proposed $\model$ offers a way
to capture the so-called ``explaining away'' effect, that is the
latent factors compete with each other as the most plausible explanatory
causes for the data (cf. also Section~\ref{sub:nrbm_intrinsic_dim}).
The competition is encouraged by the nonnegative constraints, as can
be seen from the generative model of data $p\left(\v_{n}=1\mid\bh;\psi\right)=\sigma\left(\a_{n}+\sum_{k}\w_{nk}\h_{k}\right)$,
in that some large weights (strong explaining power) will force others
to degrade or even vanish (weak explaining power). This is different
from standard practice in neural networks, where complex inhibitory
lateral connections must be introduced to model the competition \cite{hinton_ghahramani_biosci97_generative}. 

One important question is that under such constraints, besides the
obvious gains in structural regularization, do we lose representational
power of the standard RBM? On one hand, our experience has indicated
that yes, there is certain loss in the ability to reconstruct the
data, since the parameters are limited to be nonnegative. On the other
hand, we have demonstrated that this does not away translate into
the loss of predictive power. In our setting, the degree of constraints
can also be relaxed by lowering down the regularization parameter
$\alpha$ in Eq.~(\ref{eq:NRBM_reg_loglikelihood}), and this would
allow some parameters to be negative.

\subsection{Model stability}

The high-dimensional data necessitates sparse models wherein only
small numbers of strongly predictive features are selected. However,
most sparsity-inducing algorithms lead to unstable models due to data
variations (e.g., resampling by bootstrapping, slight perturbation)
\cite{xu_etal_pami12_sparse}. For example, logistic regression produces
unstable models while performing automated feature selection \cite{austin_tu_2004_automated}.
In the context of lasso, the method tends to keep only one feature
if two are highly correlated \cite{zou_hastie_jrss05_regularization},
resulting in loss of stability.

Model stability refers to the consistent degree of model parameters
which are learned under data changes. For classifiers with embedded
feature selection, a sub-problem is the stability of selected subsets
of features. The importance of stability in feature selection has
been largely ignored in literature. A popular approach in initial
work is to compare feature selection methods basing on feature preferences
ranked by their weights \cite{kvrivzek_etal_caip07_improving,kalousis_etal_kais07_stability}.
Another approach targets on developing a number of metrics to measure
stability \cite{khoshgoftaar_etal_iri13_survey}. Recently, model
stability has been studied more widely in bioinformatics. The research
focus is to improve the stability by exploiting aggregated information
\cite{park_etal_biostats07_averaged,abraham_etal_bmc10_prediction,soneson_etal_biostats12_framework}
and the redundancy in the feature set \cite{yuan_lin_jrss06_model,yu_etal_kdd08_stable}.

In this paper, our work addresses the stability problem via readmission
prognosis of heart failure patients using high-dimensional medical
records. The heart failure data is studied in \cite{he_etal_jamia13_mining}
but with only a small subset of features. Our recent work employs
graph-based regularization to stabilize linear models \cite{gopakumar_etal_jbhi14_stabilizing,tran_etal_kais14_stabilized}.
This paper differs from our previous work in that no external information
is needed. Rather, it is based on self-organization of features into
parts which are more stable than original features. This application
in healthcare analytics continues our on going research on modeling
electronic medical records which are high-dimensional and heterogeneous
data \cite{tu_etal_pakdd13_latent}.

\section{Conclusion\label{sec:conclusion}}

To summarize, this paper has introduced a novel variant of the powerful
restricted Boltzmann machine, termed \emph{nonnegative RBM} ($\model$),
where the mapping weights are constrained to be nonnegative. This
gives the $\model$ the new capacity to discover interpretable parts-based
representations, semantically plausible high-level features for additive
data such as images and texts. Our proposed method can also stabilize
linear predictive models in feature selection task for high-dimensional
medical data. In addition, the $\model$ can be used to uncover the
intrinsic dimensionality of the data, the ability not seen in the
standard RBM. This is because under the nonnegativity constraint,
the latent factors ``compete'' with each other to best represent
data, leading to some unused factors. At the same time, the $\model$
retains nearly full strength of the standard RBM, namely, compact
and discriminative distributed representation, fast inference and
incremental learning.

Compared against the well-studied parts-based decomposition scheme,
the nonnegative matrix factorization (NMF), the $\model$ could work
in places where the NMF fails. When it comes to classification using
the learned representations, the features discovered by the $\model$
are more discriminative than those by the NMF and the latent Dirichlet
allocation (LDA). For model stability, the performance of the proposed
model surpasses the lasso, the RBM and is comparable to the NMF. Thus,
we believe that the $\model$ is a fast alternative to the NMF and
LDA for a variety of data processing tasks.

\bibliographystyle{plain}
\bibliography{dami17_nrbm}

\end{document}